\documentclass{article}

\usepackage[final]{neurips_2023}

\usepackage[utf8]{inputenc} 
\usepackage[T1]{fontenc}    
\usepackage[colorlinks=true, citecolor=blue]{hyperref}       
\usepackage{url}            
\usepackage{booktabs}       
\usepackage{amsfonts}       
\usepackage{nicefrac}       
\usepackage{microtype}      
\usepackage[dvipsnames]{xcolor}         
\usepackage{lipsum}
\usepackage{import}
\usepackage{xr}

\usepackage{amsmath}
\usepackage{amssymb}
\usepackage{amsthm}
\usepackage{mathtools}
\usepackage{xcolor}
\usepackage{hyperref}
\usepackage{graphicx}
\usepackage{bm}

\usepackage{tikz}

\newlength{\indentationFormule} 
\newlength{\indentationTotaleFormule}
\newlength{\indentationCommentaire}
\newlength{\indentationDerivation}
\newlength{\largeurLangle}
\newlength{\largeurBoiteCommentaire}

{\setlength{\indentationDerivation}{#1}%
\setlength{\indentationTotaleFormule}{\indentationFormule}
\addtolength{\indentationTotaleFormule}{#1}
\setlength{\largeurBoiteCommentaire}{\linewidth}
\addtolength{\largeurBoiteCommentaire}{-\indentationFormule}
\addtolength{\largeurBoiteCommentaire}{-\indentationCommentaire}
\addtolength{\largeurBoiteCommentaire}{-\largeurLangle}
\addtolength{\largeurBoiteCommentaire}{-\indentationDerivation}
\begin{list}{}{\setlength{\leftmargin}{\indentationTotaleFormule}}
\setlength{\baselineskip}{2.3\baselineskip}
\item$}%
{\hbox{}$\end{list}}  

\newtheorem{theorem}{Theorem}
\numberwithin{theorem}{section}
\newtheorem{proposition}[theorem]{Proposition}
\numberwithin{theorem}{section}

\numberwithin{theorem}{section}
\newtheorem{lemma}[theorem]{Lemma}
\numberwithin{theorem}{section}

\theoremstyle{definition}

\numberwithin{example}{section}

\numberwithin{fact}{section}
\newtheorem{remark}{Remark}
\numberwithin{remark}{section}

\numberwithin{hypothesis}{section}
\newtheorem{definition}{Definition}
\numberwithin{definition}{section}
\newtheorem{assumption}{Assumption}

\newcommand{\somme}[3]{\sum_{#1}^{#2}{#3}}
\newcommand{\produit}[3]{\prod_{#1}^{#2}{#3}}
\newcommand{\integral}[4]{\int_{#1}^{#2} #3 \,#4}

\newcommand{\real}{\mathbb{R}}

\newcommand{\theset}[1]{\{ #1 \}}
\newcommand{\lipschitz}[1]{\text{\rm Lip}_{#1} }

\newcommand{\loss}{\ell}
\newcommand{\xfancy}{\mathcal{X}}
\newcommand{\yfancy}{\mathcal{Y}}
\newcommand{\zfancy}{\mathcal{Z}}

\newcommand{\hfancy}{\mathcal{H}}
\newcommand{\lfancy}{\mathcal{L}}
\newcommand{\ffancy}{\mathcal{F}}
\newcommand{\gfancy}{\mathcal{G}}

\newcommand{\ufancy}{\mathcal{U}}

\newcommand{\rfancy}{\mathcal{R}}

\newcommand{\zeromatrix}{\mathbf{0}}
\newcommand{\idmatrix}{\mathbf{I}}

\newcommand{\zvector}{\mathbf{z}}
\newcommand{\yvector}{\mathbf{y}}
\newcommand{\xvector}{\mathbf{x}}

\newcommand{\wvector}{\mathbf{w}}

\newcommand{\dimz}{{d_\zfancy}}

\newcommand{\equdef}{\stackrel{\text{def}}{=}}
\newcommand{\sampled}{\sim}
\newcommand{\iidsampled}{\stackrel{\text{iid}}{\sim}}
\newcommand{\norm}[1]{\left\lVert #1 \right\rVert}
\newcommand{\absolu}[1]{\left \lvert #1 \right \rvert}

\newcommand{\expon}[1]{e^{#1}}
\newcommand{\exponbig}[1]{\textnormal{exp}\left[ #1    \right]}

\newcommand{\lipnorm}[1]{\norm{#1}_{\textrm{Lip}}}

\newcommand{\pushf}[2]{{#1} \sharp #2}

\newcommand{\ipm}[1]{d_{#1}}
\newcommand{\prob}[2]{\underset{#1}{\mathbb{P}}\left[#2\right]}
\DeclareMathOperator*{\Esp}{\mathbb{E}}
\newcommand{\expect}[2]{\Esp_{#1}\left[#2\right]}
\newcommand{\expectseul}[2]{\Esp_{#1} #2 }

\newcommand{\normal}[1]{\mathcal{N}(#1)}
\newcommand{\mprob}[1]{\mathcal{M}_+^1(#1)}
\newcommand{\kl}[2]{\mathrm{KL}(#1 \, || \, #2)}

\newcommand{\abscont}{\ll}

\newcommand{\listen}[1]{#1_1, \dots, #1_n}

\newcommand{\bleu}[1]{\textcolor{blue}{#1}}
\newcommand{\details}[1]{\bleu{#1}}

\newcommand{\andspace}{\quad \text{ and } \quad}
\newcommand{\unsur}[1]{\frac{1}{#1}}
\newcommand{\omet}[1]{\details{$\bullet \bullet \bullet \bullet $}}

\newcommand{\guillemets}[1]{``#1''}

\newcommand{\muphi}[1]{\mu_\phi\left( #1  \right)}
\newcommand{\muphiseul}{\mu_\phi}
\newcommand{\sigmaphi}[1]{\sigma_\phi\left( #1  \right) }
\newcommand{\sigmaphiseul}{\sigma_\phi}
\newcommand{\sigmaphisq}[1]{\sigma_\phi^2\left( #1  \right)}
\newcommand{\gtheta}[1]{g_\theta( #1 )}
\newcommand{\gthetaseul}{g_\theta}
\newcommand{\getoil}[1]{g^*(#1)}
\newcommand{\getoilseul}{g^*}
\newcommand{\encfunc}[1]{Q_\phi \left( #1  \right)}
\newcommand{\encfuncseul}{Q_\phi}
\newcommand{\qphi}[1]{q_\phi(#1)}
\newcommand{\ptheta}[1]{p_\theta(#1)}
\newcommand{\diag}[1]{\text{diag}(#1)}

\newcommand{\ketoil}{K_*}
\newcommand{\ktheta}{K_\theta}
\newcommand{\kphi}{K_\phi}
\newcommand{\leqstar}{\stackrel{(*)}{\leq}}

\newcommand{\muhatn}{\hat{\mu}_{\phi, \theta}}
\newcommand{\otimesn}{^{\otimes n}}

\newcommand{\dimetoil}{{d^*}}
\newcommand{\petoil}{p^*}

\newcommand{\tvseul}[1]{d_{TV}}

\newcommand{\unvec}{\Vec{1}}
\newcommand{\newf}{\mathcal{E}}
\newcommand{\zetoil}{\wvector}

\newcommand{\zgivenx}{\zvector | \xvector}
\newcommand{\xgivenz}{\xvector | \zvector}
\newcommand{\cdotgivenx}{\cdot | \xvector}
\newcommand{\norme}[1]{\norm{ #1 }}  
\newcommand{\lossrec}{\loss_\text{rec}^\theta}

\newcommand{\qhatn}{\hat{q}_\phi}
\newcommand{\erf}[1]{\text{erf}\left(#1 \right)}
\newcommand{\trace}[1]{\text{Tr} \left( #1 \right)}
\newcommand{\frobnorm}[1]{\norm{#1}_{\text{Fr}}}
\newcommand{\efancy}{\mathcal{E}}
\newcommand{\lrec}{l_\text{rec}^\theta}

\makeatletter
\newcommand{\settitle}{\@maketitle}
\makeatother

\title{Statistical Guarantees for Variational Autoencoders using PAC-Bayesian Theory}

\author{%
  Sokhna Diarra Mbacke \\
  Université Laval\\
  \texttt{sokhna-diarra.mbacke.1@ulaval.ca} \\
  \And
  Florence Clerc \\
  McGill University \\
  \texttt{florence.clerc@mail.mcgill.ca} \\
  \AND
  Pascal Germain \\
  Université Laval \\
  \texttt{pascal.germain@ift.ulaval.ca} \\
}

\begin{document}

\maketitle


\begin{abstract}
Since their inception, Variational Autoencoders (VAEs) have become central in machine learning. Despite their widespread use, numerous questions regarding their theoretical properties remain open. Using PAC-Bayesian theory, this work develops statistical guarantees for VAEs. First, we derive the first PAC-Bayesian bound for posterior distributions conditioned on individual samples from the data-generating distribution. Then, we utilize this result to develop generalization guarantees for the VAE's reconstruction loss, as well as upper bounds on the distance between the input and the regenerated distributions. More importantly, we provide upper bounds on the Wasserstein distance between the input distribution and the distribution defined by the VAE's generative model. 
\end{abstract}

\section{Introduction}

In recent years, deep generative models have exhibited tremendous empirical success. Two of the most important families of generative models are Generative Adversarial Networks (GANs) \citep{gan-goodfellow} and Variational Autoencoders \citep{autoencoding, rezende-vae}. GANs take an adversarial approach, whereas VAEs are based on maximum likelihood estimation and variational inference. VAEs comprise two main components: an encoder which parameterizes an approximation of the posterior distribution over the latent variables, and a decoder which parameterizes the likelihood. In addition to generative modelling tasks such as image generation \citep{nvae} and text generation \citep{vae-text}, VAEs have been successfully applied to other topics such as semi-supervised learning \citep{vae-semi-supervised}, anomaly detection \citep{anomaly-vae}, and dimensionality reduction \citep{vae-dimensionality-reduction}. However, despite their empirical success, the question of statistical guarantees for the performance of VAEs remains largely open. Namely, how can one certify that VAEs generalize well, both in terms of reconstruction and generation?

PAC-Bayesian theory \citep{some_pb_thms,catoni} is an influential tool of statistical learning theory dedicated to providing generalization bounds for machine learning models. PAC-Bayes has been applied to a wide variety of problems such as classification \citep{pac-bayes-linear,pb-classif},  meta-learning \citep{pb-meta-learning}, co-clustering \citep{pb-co-clustering}, domain adaptation \citep{pb-domain-adaptation}, and online learning \citep{online-pb}. In recent years, PAC-Bayes has been used to derive non-vacuous generalization bounds for supervised learning algorithms based on neural networks \citep{diff-privacy, perez-ortiz}. See \citet{primer} and \citet{friendly} for excellent surveys.

The objective of this work is to utilize PAC-Bayesian theory to derive statistical guarantees for VAEs. Our generalization bounds investigate the reconstruction, regeneration, as well as the generation properties of VAEs.

\subsection{Related Works} 

 In order to explain the empirical success of deep generative models, a lot of attention has been put into deriving theoretical guarantees for these models. Most of the results, however, have been dedicated to GANs and their variants \citep{gen-equilibrium, disc-gen-tradeoff, liang, singh, without-domination, insights, pb-gen-models}. A possible explanation for this plethora of theoretical results is the adversarial loss function, which directly offers an estimation of the discrepancy between the input distribution and the generator's distribution.
Despite being central tools in modern machine learning, VAEs have not benefited from such a thorough theoretical analysis \citep{regeneration}.

The work of \citet{regeneration} studies the regeneration properties of Wasserstein autoencoders (WAEs) \citep{wae}, which come from the same family as VAEs. Using VC theory, \citet{regeneration} derive rates of convergence for the Wasserstein distance between the input distribution and the distribution regenerated by the WAE, as well as the total variation distance between the empirical latent distribution and the latent prior.  
Taking a more empirical approach, \citet{pb-vae} use PAC-Bayes to study the generalization properties of stochastic reconstruction models. They define a $[0, 1]$-bounded reconstruction loss function, then utilize McAllester's bound \citep{mc2003a} to formulate a generalization bound for models with probabilistic neural networks \citep{langford2001}. Then, they re-scale their loss and compare the empirical results to the reconstruction of standard VAEs on benchmark datasets.

We also mention the work of \citet{pb-gen-models}, who developed PAC-Bayesian bounds for the analysis of adversarial generative models. Using McDiarmid's inequality, they proved upper bounds on the distance between the input distribution and the generator's distribution, for WGANs \citep{wgan} and EBGANs \citep{ebgan}. 

\subsection{Our Contributions}
In this work, we derive theoretical guarantees for variational autoencoders using PAC-Bayesian theory. We provide three types of guarantees: reconstruction guarantees showing that VAEs can successfully reconstruct unseen samples from the input distribution; regeneration guarantees proving upper bounds on the Wasserstein distance between the input distribution and the distribution regenerated by the VAE, given the training set as input; and finally, generation guarantees showing upper bounds on the Wasserstein distance between the data-generating distribution and the VAE's generated distribution defined by the latent prior and the decoder. To the best of our knowledge, these are the first generalization bounds for the standard VAE's reconstruction and regeneration properties, as well as the first statistical guarantees for the VAE's generative model.

In our analysis, the PAC-Bayesian posterior coincides with the variational posterior, which requires the PAC-Bayesian posterior to be conditional. Since, to the best of our knowledge, such PAC-Bayes bounds do not exist in the literature, we start by developing the first PAC-Bayesian bound for conditional posterior distributions. Then, we provide upper bounds for the VAE's performance under two main assumptions: we start by assuming the instance space is bounded, then we take advantage of the manifold hypothesis. Our bounds are functions of the optimization objective of the VAE, namely, the empirical reconstruction loss, and the empirical KL-loss. 

The remainder of this paper is organized as follows. In Section \ref{sec-basics}, we define some preliminary concepts, then briefly introduce VAEs and PAC-Bayesian theory. Section \ref{sec-pb-bound} presents our general PAC-Bayesian theorem for conditional posteriors. Then, in  Sections \ref{sec-recons} and \ref{sec-gen}, we present our generalization bounds for the reconstruction loss, and the regeneration and generation guarantees.

\section{Preliminaries} \label{sec-basics}

\subsection{Definitions and Notations}
Given metric spaces $(\xfancy, d)$ and $(\yfancy, d')$, and a real number $K>0$, a function $f:\xfancy \rightarrow \yfancy$ is $K$-Lipschitz continuous if for any $\xvector, \yvector \in \xfancy$, we have
\[
d'(f(\xvector), f(\yvector)) \leq K d(\xvector, \yvector).
\]
The smallest $K$ such that this condition is satisfied is called the \emph{Lipschitz norm} or \emph{Lipschitz constant} of $f$ and is denoted $\lipnorm{f}$. Moreover, the set of $K$-Lipschitz continuous functions $f:\xfancy \rightarrow \yfancy$ is denoted $\lipschitz{K}(\xfancy, \yfancy)$ (the underlying metrics will be clear from the context).

Throughout the paper, we use lower case letters $p, q$ to denote both probability distributions and their densities w.r.t. the Lebesgue measure. We may add variables between parentheses to improve readability (e.g. $p(\zvector)$ to emphasize that $p$ is a distribution on the space of variables $\zvector$, and $q(\zvector|\xvector)$ to indicate that $q$ is a conditional distribution). The set of probability measures on a space $\xfancy$ is denoted $\mprob{\xfancy}$. The Kullback–Leibler (KL) divergence between $p, q \in \mprob{\xfancy}$ is denoted $\kl{p}{q}$. We omit the absolute continuity condition $p \abscont q$ in the statements of the results below, since if it is not satisfied, then one may assume the KL divergence is infinite and the bounds hold trivially. 

Integral Probability Metrics (IPM, see \citet{ipm-def}) are a class of pseudo-metrics defined on the space of probability measures. Given a family $\ffancy$ of real-valued functions defined on $\xfancy$, the IPM defined by $\ffancy$ is denoted $\ipm{\ffancy}$ and defined as
\begin{equation} \label{eq-def-ipm}
\ipm{\ffancy}(p, q) = \sup_{f\in\ffancy}\absolu{ \integral{}{}{f}{dp} - \integral{}{}{f}{dq}}, \quad \forall p, q \in \mprob{\xfancy}. 
\end{equation}

Stemming from the theory of optimal transportation \citep{villani}, the Wasserstein distances (see Definition~\ref{def-app-wk}) are a class of metrics between probability measures. The Wasserstein distance of order $1$, also referred to simply as \emph{the Wasserstein distance}, is the IPM defined by the set $\ffancy = \theset{f: \xfancy \rightarrow \real \text{ s.t. } \lipnorm{f} \leq 1}$.

Finally, we recall the definition of a \emph{pushforward measure}. Let $p$ be a probability distribution on a space $\zfancy$ and $g: \zfancy \rightarrow \xfancy$ be a measurable function. The pushforward measure defined by $g$ and $p$ and denoted $\pushf{g}{p}$ is a probability distribution on $\xfancy$ defined as $\pushf{g}{p}(A) = p(g^{-1}(A))$, for any measurable set $A \subseteq \xfancy$. In other words, sampling $\xvector \sampled \pushf{g}{p}$ means sampling $\zvector \sampled p$ first, then setting $\xvector=g(\zvector)$.

\subsection{Variational Autoencoders}
We consider a Euclidean observation space $\xfancy$, a data-generating distribution $\mu \in \mprob{\xfancy}$, and a latent space $\zfancy=\real^\dimz$. 
VAEs comprise two main components: the encoder network whose parameters are denoted $\phi$, and the decoder network whose parameters are denoted $\theta$. For simplicity, we may refer to $\phi$ and $\theta$ as the encoder and decoder respectively. The encoder parameterizes a distribution $\qphi{\zgivenx}$ over the latent space $\zfancy$, which is a variational approximation of the Bayesian posterior $\ptheta{\zgivenx}$. The likelihood $\ptheta{\xgivenz}$ is parameterized by the decoder network. In this work, we consider the standard VAE, with a standard Gaussian prior $p(\zvector)=\normal{\zeromatrix, \idmatrix}$ on $\zfancy$ and Gaussian latent distributions $\qphi{\zgivenx}$. 
More precisely, for any $\xvector \in \xfancy$, the distribution $\qphi{\zgivenx}$ is a Gaussian distribution with a diagonal covariance matrix $\normal{\muphi{\xvector}, \diag{\sigmaphisq{\xvector}}}$, where
\[
\muphiseul: \xfancy \rightarrow \zfancy = \real^\dimz       \andspace     \sigmaphiseul : \xfancy \rightarrow \real^\dimz_{\geq 0}.
\]
Note that $\diag{\sigma}$ denotes the diagonal matrix whose main diagonal is the vector $\sigma$. In order to simplify some of the expressions below, it may be useful to express the encoder network as a function
\begin{equation} \label{eq-qphi}
\encfuncseul : \xfancy \rightarrow \real^{2\dimz}, \quad \text{where } 
\encfunc{\xvector} = \begin{bmatrix}
   \muphi{\xvector} \\
   \sigmaphi{\xvector}
 \end{bmatrix}.
\end{equation}
We express the decoder as a parametric function $\gthetaseul: \zfancy \rightarrow \xfancy$. For any $\xvector \in \xfancy$, upon receiving $\zvector \sampled \qphi{\zgivenx}$, the decoder's output $\gtheta{\zvector}$ is a reconstruction of $\xvector$. Given a training set $S = \theset{\listen{\xvector}}$, the encoder and decoder networks are jointly trained by minimizing the following objective: 
\begin{equation}\label{eq-loss-vae}
\lfancy_\text{VAE}(\phi,\theta) = \unsur{n} \somme{i=1}{n}{  \left[   \expect{\zvector \sampled \qphi{\zgivenx_i}}{-\log \ptheta{\xvector_i | \zvector}} + \beta \kl{\qphi{\zgivenx_i}}{p(\zvector)}   \right]  },
\end{equation}
where the first part of \eqref{eq-loss-vae} is the \emph{reconstruction loss} and the second part is the KL-divergence between the latent distributions (associated to the training samples) and the prior over the latent space, weighted by a hyperparameter $\beta > 0$ \citep{beta-vae}. 
The reconstruction loss measures the similarity between $\xvector$ and its reconstruction $ \gtheta{\zvector}$, and can be defined in many ways. With a Gaussian likelihood, the reconstruction loss is the squared $L_2$ norm $\norme{\xvector - \gtheta{\zvector}}^2  $.

After training, the VAE defines a generative model using the prior $p(\zvector)$ and the decoder $\gthetaseul$ \citep{autoencoding}. The distribution $\pushf{\gthetaseul}{p(\zvector)} \in \mprob{\xfancy}$ allows one to generate new samples by first sampling a latent vector from the prior, then passing it through the decoder. We refer to $\pushf{\gthetaseul}{p(\zvector)}$ as the VAE's generated distribution.

\subsection{A Brief Introduction to PAC-Bayesian Theory}
Dating back to \citet{some_pb_thms}, PAC-Bayesian theory develops high-probability generalization bounds for machine learning algorithms. In essence, PAC-Bayes frames the output of such algorithm as a posterior distribution over a class of hypotheses, and provides an upper bound on the discrepancy between a model's empirical risk and its population risk. 

PAC-Bayes considers the following concepts: a hypothesis class $\hfancy$, a training set $S=\theset{\listen{\xvector}}$ iid sampled from an unknown distribution $\mu$ over an instance space $\xfancy$\footnote{In supervised learning, the instance space has the form $\xfancy \times \yfancy$ where $\xfancy$ is a set of features, and $\yfancy$ a set of labels. We use a more general formulation to encompass the unsupervised learning setting.}, and a real-valued loss function $\loss : \hfancy \times \xfancy \rightarrow [0, \infty)$. Moreover, the primary goal of PAC-Bayes is to provide generalization bounds uniformly valid for any posterior $q\in \mprob{\hfancy}$. These bounds are dependent on the empirical performance of $q$ and its closeness to a chosen \emph{prior distribution} $p \in \mprob{\hfancy}$, as measured by the KL-divergence. The empirical and true risks of a posterior distribution $q \in \mprob{\hfancy}$ are defined as 
\[
\hat{\rfancy}_S(q) = \expect{h \sampled q(h)}{\unsur{n}\somme{i=1}{n}{ \loss(h, \xvector_i) }} \andspace
\rfancy(q) = \expect{h \sampled q(h)}{\expectseul{\xvector\sampled \mu}{\loss(h, \xvector)}}.
\]
As an illustration, consider the following PAC-Bayesian bound for bounded loss functions developed by \citet{catoni}.
\begin{theorem}
Given a probability measure $\mu$ on $\xfancy$, a hypothesis class $\hfancy$, a prior distribution $p$ on $\hfancy$, a loss function $\loss: \hfancy \times \xfancy \rightarrow [0, 1]$, real numbers $\delta \in (0, 1)$ and $\lambda >0$, with probability at least $1-\delta$ over the random draw of $S\sampled \mu\otimesn$, the following holds for any posterior $q\in \mprob{\hfancy}$:
\[
\rfancy(q) \leq \hat{\rfancy}_S(q) + \frac{\lambda}{8n} + \frac{\kl{q}{p} + \log\unsur{\delta}}{\lambda}.
\]
\end{theorem}

The connection between PAC-Bayesian theory and Bayesian inference was highlighted by \citet{Grunwald12} and \citet{meets}, who showed that with a proper choice of $\lambda$ and the negative log-likelihood as the loss function $\loss$, the optimal posterior minimizing the right-hand side of Catoni's bound is the Bayesian posterior.
Note that although the Bayesian posterior is unique (for a given prior and likelihood), a \guillemets{PAC-Bayesian posterior} could be, in principle, any distribution over $\hfancy$.

In our PAC-Bayesian analysis of VAEs, we will use the latent space $\zfancy$ as our hypothesis class, so that the VAE's prior will coincide with the PAC-Bayesian prior and the variational posterior $\qphi{\zgivenx}$ will stand for our PAC-Bayesian posterior. An immediate concern with this approach is that the encoder's distributions are conditioned on individual samples $\xvector\sampled \mu$, whereas the usual PAC-Bayesian bounds hold for unconditional posteriors $q(h)$.
We address this issue in the next section, by developing a novel PAC-Bayesian bound for posterior distributions $q(\cdotgivenx)$. This general result will be later utilized to analyze VAEs.

\section{A General PAC-Bayesian Bound with a Conditional Posterior}\label{sec-pb-bound}

In this section, we present our general PAC-Bayesian bound with a conditional posterior distribution. Note that the novelty of this result is not the conditioning on observations, since this can be achieved by exploiting the existing PAC-Bayesian bounds. Indeed, \citet{online-pb} utilized the general theorem of \citet{beyond-usual} to derive bounds for the online learning framework. Instead, the contribution of Theorem \ref{thm-after-ipm-assum} is to 
predict the behavior of $q(h|\xvector)$, for any (previously unseen) $\xvector \sampled \mu$, when the posterior $q$ was only learned using the training samples $\theset{\listen{\xvector}}$.
To the best of our knowledge, this is the first PAC-Bayesian bound where the posterior distribution is a conditional distribution conditioned on individual elements from the instance space. This bound will require the posterior $q$ and the loss function $\loss$ to satisfy the following technical assumption.

\begin{assumption}\label{assum-ipm}

We say that a distribution $q(\cdotgivenx)$ and a loss function $\loss$ satisfy Assumption 1 with a constant $K > 0$ if there exists a family $\newf$ of functions $\hfancy \rightarrow \real$ such that the following properties hold.
\begin{enumerate}
\item The function $\xvector \mapsto q(\cdotgivenx)$ is continuous in the following sense:
for any $\xvector_1, \xvector_2 \in \xfancy$,
\[
\ipm{\newf}\left( q(h|\xvector_1), q(h|\xvector_2)  \right) \leq K d(\xvector_1 , \xvector_2).
\]

\item For any $\xvector\in \xfancy$, the function $\loss(\cdot, \xvector) : \hfancy \rightarrow \real$ is in $\newf $:
\[
\loss(\cdot, \xvector) \in \newf, \quad \text{for any } \xvector\in \xfancy.
\]
\end{enumerate}
\end{assumption}

Before stating the general result, let us pause and discuss this assumption. Intuitively, the goal of a generalization bound is to predict the behavior of the posterior distribution $q(h|\xvector)$ on previously unseen examples $\xvector\sampled \mu$. Since the posterior $q(h|\xvector)$ is learned by minimizing the loss function $\loss$ on the training samples $S= \theset{\listen{\xvector}}$, one may need two things to be true. 

First, the mapping $\xvector \mapsto q(h|\xvector)$
has to be somewhat continuous. This is ensured by the first part of Assumption \ref{assum-ipm}, which states that the posterior $q$ is Lipschitz-continuous\footnote{$\ipm{\newf}$ is a pseudo-metric in the general case, so we abuse the definition by calling this Lipschitz continuity, since the latter concept is only defined for metric spaces.} with respect to the IPM $\ipm{\newf}$ and the underlying metric $d$ on $\xfancy$. Indeed, this tells us that if $\xvector_1$ and $\xvector_2$ are close w.r.t. the underlying metric on $\xfancy$, then $q(h|\xvector_1)$ and $q(h|\xvector_2)$ are close, w.r.t. the IPM $\ipm{\newf}$.

Second, that continuity has to be \guillemets{understood} by the loss function $\loss$, which corresponds to the second part of the assumption. It states that the loss function's discriminative power is weaker than the one defined by the IPM $\ipm{\newf}$. In other words, the discrepancy measure used to measure the similarity between the distributions $q(h|\xvector_1)$ and $q(h|\xvector_2)$ needs to be just strong enough to fool the loss function into thinking that the distributions are close to each other. An alternate formulation of Assumption~\ref{assum-ipm} is provided in the supplementary material (Remark~\ref{rem-assum-alternate}).

Finally, we emphasize that Assumption \ref{assum-ipm} is not as restrictive as it may seem at first. For instance, it is satisfied by a VAE's variational posterior, when the encoder and decoder networks have finite Lipschitz norms and the reconstruction loss is defined with the $L_2$ norm (see Proposition \ref{prop-assum-ipm}).
We are ready to state our first result.
\begin{theorem}\label{thm-after-ipm-assum}
Let $(\xfancy, d)$ be a metric space. Consider a probability measure $\mu$ on $\xfancy$, a hypothesis class $\hfancy$, a prior distribution $p(h)$ on $\hfancy$, a loss function $\loss: \hfancy {\times} \xfancy {\rightarrow} \real$, real numbers $\delta \in (0, 1)$ and $\lambda \,{>}\, 0$. With probability at least $1\,{-}\,\delta$ over the random draw of $S \,{\sampled}\, \mu\otimesn$, the following holds for any conditional posterior $q(h|\xvector)$ such that Assumption \ref{assum-ipm} is satisfied by $q(h|\xvector)$ and $\loss$ with constant $K \,{>}\, 0$:
\begin{equation*} 
\begin{split}
\expectseul{\xvector\sampled \mu}{\expectseul{h\sampled q(h|\xvector)}{\hspace{-3mm}\loss(h, \xvector)}} - \unsur{n}\somme{i=1}{n}{\expectseul{h\sampled q(h|\xvector_i)}{\hspace{-4mm}\loss(h, \xvector_i)}}
   \leq
\unsur{\lambda} \left[  
\somme{i=1}{n}{\kl{q(h|\xvector_i)}{p(h)}} +
\frac{\lambda K}{n} \somme{i=1}{n}{\expectseul{\xvector\sampled \mu}{d(\xvector, \xvector_i)}} + {}
\right. \\  \left.
\log \unsur{\delta} + n \log \expectseul{h \sampled p(h)}{\expectseul{\xvector \sampled \mu}{\expon{\frac{\lambda}{n} \left( \expectseul{\xvector' \sampled \mu}{\loss(h, \xvector')} - \loss(h, \xvector) \right) }}}
  \right].
\end{split}
\end{equation*}
\end{theorem}
In order to prove Theorem \ref{thm-after-ipm-assum}, we start by deriving a bound where the expected loss for samples $\xvector\sampled \mu$  is computed w.r.t. distributions $q(h|\xvector_i)$ associated to the training samples (see Lemma~\ref{lem-before-ipm-assum}). This result uses standard PAC-Bayesian techniques, with a key difference: we start with $n$ iid hypotheses from the prior $p(h)$, then we perform the change of measure with $n$ posteriors $\qphi{\zgivenx_1}, \dots, \qphi{\zgivenx_n}$, and show that the resulting exponential moment is equal to the one in Theorem \ref{thm-after-ipm-assum}.
Moreover, one of the original aspects of this work comes from Assumption \ref{assum-ipm}, which enables us to obtain a bound where the expected loss for $\xvector\sampled \mu$ is computed w.r.t. the posterior $q(h|\xvector)$, associated to $\xvector$ itself instead of all the training samples. However, the price to pay for having a posterior $q(h|\xvector)$ for each $\xvector\in \xfancy$ is that the bound depends on  $\frac{1}{n}\somme{i=1}{n}{\expectseul{\xvector\sampled \mu}{d(\xvector, \xvector_i)}}$, which we refer to as the \emph{average distance}.

Applied to supervised learning, Theorem \ref{thm-after-ipm-assum} bounds the expected risk of a Gibbs posterior $q$ which, upon receiving a previously unseen datapoint $\xvector \sampled \mu$, samples a predictor $h$ \emph{dependent} on $\xvector$, and uses it to make a prediction. Note that the family $\newf$ from Assumption \ref{assum-ipm} does not appear in the bound, which has nice consequences in practice. Indeed one may pick a loss function $\loss$ that fits the problem, and then find a family $\newf$ for which the continuity assumption is satisfied with constant $K$ that is as small as possible.

Note also that, in the tradition of PAC-Bayesian bounds, Theorem \ref{thm-after-ipm-assum} does not make any assumptions on the nature of the elements of $\hfancy$ (e.g. $\hfancy$ could be a class of functions, a set of neural network's parameters, etc). Therefore, the theorem is very general and could be applied to different domains and models. In the following sections, we will use a specific kind of hypothesis class $\hfancy = \zfancy$, in order to capture the VAE's latent space.


\section{Generalization bounds for the Reconstruction Loss}\label{sec-recons}

For the remainder of this work, $\norm{\cdot}$ denotes the $L_2$ norm, and we assume the instance space $\xfancy$ is Euclidean, and the latent space $\zfancy = \real^\dimz$, where $\dimz > 0$. Both $\xfancy$ and $\zfancy$ are equipped with the Euclidean distance as the underlying metric. Therefore, if $\xvector, \xvector' \in \xfancy$, $d(\xvector, \xvector') = \norm{\xvector - \xvector'}$.

The following assumption states that the encoder and decoder networks have finite Lipschitz norms.
\begin{assumption} \label{assum-lipschitz}
The encoder and decoder are Lipschitz-continuous w.r.t. their inputs, meaning there exist real numbers $\kphi, \ktheta > 0$ such that for any $\xvector_1, \xvector_2 \in \xfancy$ and $\zvector_1, \zvector_2 \in \zfancy$,
\begin{equation}\label{eq-kphi}
\norme{\encfunc{\xvector_1} - \encfunc{\xvector_2}} \leq \kphi \norme{\xvector_1 - \xvector_2}
\end{equation}
and 
\begin{equation}\label{eq-ktheta}
\norme{\gtheta{\zvector_1} - \gtheta{\zvector_2}} \leq \ktheta \norme{\zvector_1 - \zvector_2}.
\end{equation}
Recall the definition of $\encfuncseul$ from Equation~\eqref{eq-qphi}. Note that in practice, one can estimate the Lipschitz constant of trained networks \citep{lip-efficient, lip-constant-estimation} or train the VAE with preset Lipschitz constants \citep{cert-robust-vaes}.
\end{assumption}
Moreover, we define the reconstruction loss $\lossrec$ with the $L_2$ norm, instead of the squared $L_2$ norm, which enables us to exploit the properties of a metric. We discuss this choice in Section~\ref{sec-discuss}. In order to be consistent with the PAC-Bayesian framework, we define the loss function as follows: $\lossrec : \zfancy \times \xfancy \rightarrow [0, \infty)$,
\begin{equation}\label{eq-lossrec}
\lossrec(\zvector, \xvector) = \norme{\xvector - \gtheta{\zvector}}.
\end{equation}

Our goal is to apply the general bound of Theorem \ref{thm-after-ipm-assum} to the VAE model. But first, since Theorem \ref{thm-after-ipm-assum} requires Assumption \ref{assum-ipm} to be satisfied, we start by showing that if the encoder and decoder networks have finite Lipschitz norms, then Assumption \ref{assum-ipm} holds.

\begin{proposition}\label{prop-assum-ipm}
Consider a VAE with parameters $\phi$ and $\theta$ and let $\kphi, \ktheta\in \real$ be the Lipschitz norms of the encoder and decoder respectively. Then the variational distribution $\qphi{\zgivenx}$ satisfies Assumption \ref{assum-ipm}, with $\newf = \theset{f: \zfancy \rightarrow \real \,\, s.t. \lipnorm{f} \leq \ktheta}$, $\loss=\lossrec$, and $K=\kphi \ktheta$.
\end{proposition}

\begin{proof}[Proof idea]
The proof of Proposition~\ref{prop-assum-ipm} is in Appendix \ref{sec-app-recons}, we provide a brief summary here. To prove the first part of Assumption \ref{assum-ipm}, we first notice that if $\newf$ is the set of real-valued $\ktheta$-Lipschitz continuous functions, then $\ipm{\newf}$ is a scaling of the Wasserstein distance. In addition, since $W_1 \leq W_2$, using the closed form of the Wasserstein-2 distance between Gaussian distributions, one can show that $\ipm{\newf}(\qphi{\cdotgivenx_1}, \qphi{\cdotgivenx_2}) \leq \kphi \ktheta \norme{\xvector_1 - \xvector_2}$. Finally, the second part of the assumption is a consequence of the definition of the loss function and the Lipschitz continuity of the decoder.
\end{proof}

Proposition \ref{prop-assum-ipm} tells us that Assumption \ref{assum-ipm} holds for VAEs. Consequently, we can utilize our general bound of Theorem~\ref{thm-after-ipm-assum} to obtain generalization guarantees. This leads to the following general PAC-Bayesian bound for the VAE's reconstruction loss.

\begin{theorem} \label{thm-gen-vae}
Let $\xfancy$ be the instance space, $\mu \in \mprob{\xfancy}$ the data-generating distribution, $\zfancy$ the latent space, $p(\zvector) \in \mprob{\zfancy}$ the prior distribution on the latent space, $\theta$ the decoder's parameters,  $\delta \in (0, 1), \lambda > 0$ be real numbers. With probability at least $1-\delta$ over the random draw of $S\sampled \mu\otimesn$, the following holds for any posterior $\qphi{\zgivenx}$:
\begin{equation*} 
\begin{split}
\expectseul{\xvector\sampled \mu}&{ \,\, \expectseul{ \qphi{\zgivenx}}{\lossrec(\zvector, \xvector)}} \leq \unsur{n}\somme{i=1}{n}{\expectseul{\qphi{\zgivenx_i}}{\lossrec(\zvector, \xvector_i)}}
   +
\unsur{\lambda} \left[
\somme{i=1}{n}{\kl{\qphi{\zgivenx_i}}{p(\zvector)}} + \right. \\  
&\left.
\tfrac{\lambda \kphi\ktheta}{n} \somme{i=1}{n}{\expectseul{\xvector\sampled \mu}{d(\xvector, \xvector_i)}} + 
\log \tfrac{1}{\delta} + n \log \expectseul{\zvector \sampled p(\zvector)}{\expectseul{\xvector \sampled \mu}{\expon{\frac{\lambda}{n} \left( \expectseul{\xvector' \sampled \mu}{\lossrec(\zvector, \xvector')} - \lossrec(\zvector, \xvector) \right) }}}  \right],
\end{split}
\end{equation*}
where $\kphi$ and $\ktheta$ are the Lipschitz norms of the encoder and the decoder 
(see \eqref{eq-kphi} and~\eqref{eq-ktheta}) and $\expectseul{\qphi{\zgivenx}}$ is a shorthand for $\expectseul{\zvector\sampled \qphi{\zgivenx}}$.

\end{theorem}

Note that the choice of the hyperparameter $\beta$ in the VAE's optimization objective \eqref{eq-loss-vae} correlates with the choice of the hyperparameter $\lambda$ in Theorem \ref{thm-gen-vae} (e.g. $\lambda =n$ corresponds to $\beta =1$). 
Note also that the encoder and decoder are not treated the same way in Theorem \ref{thm-gen-vae}. Indeed, the inequality holds for a given decoder, but uniformly for any encoder. We discuss this subtle difference and its practical consequences in Section~\ref{sec-discuss}.

Theorem \ref{thm-gen-vae} can be seen as a general framework. In order to obtain a useful upper bound, one needs to bound the average distance and the exponential moment on the right-hand side. In the sections below, we provide upper bounds for these terms under various assumptions on the instance space.

\subsection{Reconstruction Guarantees for Bounded Instance Spaces}
In the following theorem, we provide a special case of Theorem \ref{thm-gen-vae} when the instance space's diameter $\Delta \equdef \sup_{\xvector, \xvector' \in \xfancy}d(\xvector, \xvector')$ is finite (see Section \ref{sec-proof-thm-rec-bounded} for the proof).

\begin{theorem}\label{thm-rec-bounded}
Let $\xfancy$ be the instance space, $\Delta <\infty$ its diameter, $\mu \in \mprob{\xfancy}$ the data-generating distribution, $\zfancy$ the latent space, $p(\zvector) \in \mprob{\zfancy}$ the prior on the latent space, $\theta$ the decoder's parameters, $\delta \in (0, 1), \lambda > 0$ be real numbers. With probability at least $1-\delta$ over the random draw of $S\sampled \mu\otimesn$, the following holds for any posterior $\qphi{\zgivenx}$:
\begin{equation*}
\begin{split}
 \expectseul{\xvector\sampled \mu }{\expectseul{\qphi{\zvector|\xvector}  }{\lossrec(\zvector, \xvector)}}      \leq
 \frac{1}{n} \somme{i=1}{n}{ \left\{ \expectseul{ \qphi{\zvector|\xvector_i}   }{\lossrec(\zvector, \xvector_i)}  \right\}    }  
  + \unsur{\lambda} \left( 
  \somme{i=1}{n}{\kl{\qphi{\zvector|\xvector_i}}{p(\zvector)}  }  +{} \right. \\
  \lambda \kphi \ktheta \Delta  +  \left.
\log \unsur{\delta} + \frac{\lambda^2 \Delta^2}{8n} \right).
\end{split}
\end{equation*}
\end{theorem}
The left-hand side of this inequality is the expected reconstruction loss for samples $\xvector\sampled \mu$, while the right-hand side is the empirical reconstruction and KL losses, plus an additional term depending on the Lipschitz constants of the VAE and the model's diameter. 

Note that for real-life datasets, the diameter of the instance space might be very large and non-representative of the structure and complexity of the data. Indeed, it is common to scale image datasets in order to utilize a specific architecture \citep{dcgan}. 
In the following section, we provide a special case of Theorem \ref{thm-gen-vae} under the manifold hypothesis on the data-generating process.

\subsection{Reconstruction Guarantees under the Manifold Assumption} \label{sec-recons-manif}

The manifold assumption \citep{survey-dimension-reduction, manifold-hypo, test-manifold-1} states that most high-dimensional datasets encountered in practice lie close to low-dimensional manifolds. 
This assumption is exploited by latent variable generative models such as GANs and VAEs, which approximate high-dimensional datasets using transformations of distributions on a low-dimensional space. The works of \citet{without-domination} and \citet{pb-gen-models} provide generalization bounds for GANs,
by assuming that the data-generating distribution is a smooth transformation of the uniform distribution on $[0, 1]^\dimetoil$, where $\dimetoil$ is the intrinsic dimension. However, since the standard VAE calls for a standard Gaussian prior, in the following theorem, we assume $\mu$ is a smooth transformation of the standard Gaussian distribution $\petoil$ on $\real^\dimetoil$. We consider the case when $\petoil$ is the uniform distribution on $[0, 1]^\dimetoil$ in the supplementary material.

\begin{theorem}\label{thm-rec-manif}
Let $\xfancy$ be the instance space, $\mu \in \mprob{\xfancy}$ the data-generating distribution, $\zfancy$ the latent space, $p(\zvector) \in \mprob{\zfancy}$ the prior distribution on the latent space, $\theta$ the decoder's parameters, $\delta \in (0, 1), \lambda >0, a>0$  real numbers. Assume the data-generating distribution $\mu = \pushf{\getoilseul}{\petoil}$, where $\petoil$ is the standard Gaussian distribution on $\real^\dimetoil$ and $\getoilseul \in \lipschitz{\ketoil}(\real^\dimetoil, \xfancy)$.
With probability at least $1 - \delta - \frac{n \dimetoil}{2} e^{-a^2 / 2}$ over the random draw of $S \sampled \mu\otimesn$, the following holds for any posterior $\qphi{\zgivenx}$:
\begin{equation*}
\begin{split}
 \expectseul{\xvector\sampled \mu }{\expectseul{\qphi{\zvector|\xvector}  }{\lossrec(\zvector, \xvector)}}      \leq
 \frac{1}{n} \somme{i=1}{n}{ \left\{ \expectseul{ \qphi{\zvector|\xvector_i}   }{\lossrec(\zvector, \xvector_i)}  \right\}    }  
  + \unsur{\lambda} \left( 
  \somme{i=1}{n}{\kl{\qphi{\zvector|\xvector_i}}{p(\zvector)}  }  +{} \right. \\
   \lambda \kphi \ktheta \ketoil \sqrt{(1 + a^2)\dimetoil}  +  \left.
\log \unsur{\delta} + \frac{\lambda^2 \ketoil^2}{2n} \right).
\end{split}
\end{equation*}
\end{theorem}
Let us clarify the role of the new parameter $a > 0$. Each training sample $\xvector_i \in S$ can be expressed as $\xvector_i = \getoil{\zetoil_i}$, where $\zetoil_i \sampled \petoil$. Since $\petoil$ is the standard Gaussian distribution on $\real^\dimetoil$, all samples $\zetoil_i$ will be inside a hypercube $[-a, a]^\dimetoil$, with high probability. This uncertainty is reflected in the lowered confidence (from $1-\delta$ in Theorem \ref{thm-gen-vae} to $1-\delta - \frac{n\dimetoil}{2}\expon{-a^2/2}$ in Theorem \ref{thm-rec-manif}), and can be controlled by choosing a large enough value of $a$. The proof of Theorem \ref{thm-rec-manif} is in the supplementary material (Section~\ref{sec-app-rec-manif}), we provide a short summary below.

\begin{proof}[Proof idea]
The proof starts with Theorem~\ref{thm-gen-vae}, and uses the assumptions of Theorem~\ref{thm-rec-manif} to obtain upper bounds on the exponential moment and the average distance. To derive the upper bound on the exponential moment, we observe that the function $\zvector \mapsto \lossrec(\zvector, \xvector)$ is $\ketoil$-Lipschitz continuous, then we use a dimension-free upper bound on the MGF of Lipschitz-continuous functions of Gaussian random variables. Furthermore, we obtain the upper bound on the average distance $ \unsur{n}\somme{i=1}{n}{\expectseul{\xvector\sampled \mu}{\norme{\xvector - \xvector_i}}}$, by using Holder's inequality and the expectation of a non-central $\chi^2$ distribution. Then, we upper-bound the probability that $\zetoil_i \in [-a, a]^\dimetoil$ for all $1\leq i \leq n$ using the error function and Bernoulli's inequality. Finally, we use the union bound to update the overall confidence.
\end{proof}

\section{Generalization Bounds for Regeneration and Generation}\label{sec-gen}

Let $\muhatn$ be the \emph{empirical regenerated distribution}, meaning
\begin{equation}
\muhatn = \textstyle\unsur{n} \somme{i=1}{n}{\pushf{\gthetaseul}{\qphi{\zgivenx_i}}}.
\end{equation}

In other words, sampling $\xvector \sampled \muhatn$ is done by sampling $\zvector \sampled \qphi{\zgivenx_i}$ where $i$ is uniformly sampled from $\theset{1, \dots, n}$, then passing $\zvector$ through the decoder: $\xvector=\gtheta{\zvector}$. It is therefore the distribution regenerated by the VAE, given the training set $S = \theset{\listen{\xvector}}$ as input. 

In this section, we provide statistical guarantees on the regenerative and generative properties of VAEs. More precisely, we derive upper bounds for the quantities $W_1(\mu, \muhatn)$ and $W_1(\mu, \pushf{\gthetaseul}{p(\zvector)})$. Note that the average distance term does not appear in the bounds of this section. This is because instead of relying on Theorem~\ref{thm-after-ipm-assum}, the results of this section depend upon a preliminary lemma (Lemma~\ref{lem-before-ipm-assum}), which does not necessitate Assumption~\ref{assum-ipm}.

\subsection{Regeneration and Generation Guarantees for Bounded Instance Spaces}
The following theorem presents our first upper bound on the distance between the input distribution and the empirical regenerated distribution.  
\begin{theorem}\label{thm-regen-bounded}
Under the definitions and assumptions of Theorem \ref{thm-rec-bounded}, we have that with probability at least $1-\delta$ over the random draw of $S\sampled \mu\otimesn$, the following holds for any posterior $\qphi{\zgivenx}$:
\begin{equation*}
\begin{split}
W_1(\mu, \muhatn) \leq 
\frac{1}{n} \somme{i=1}{n}{ \left\{ \expectseul{ \qphi{\zvector|\xvector_i}   }{\lossrec(\zvector, \xvector_i)}  \right\}    }  
  + 
\unsur{\lambda} \left( \somme{i=1}{n}{\kl{\qphi{\zvector|\xvector_i}}{p(\zvector)}  }  + 
\log \unsur{\delta} + \frac{\lambda^2 \Delta^2}{8n}   \right).
\end{split}
\end{equation*}
\end{theorem}

As we can see, the right-hand side of Theorem \ref{thm-regen-bounded} depends on the empirical reconstruction loss and KL-divergence. This guarantees that as the VAE's empirical risk decreases, the regenerated distribution gets closer to the data-generating distribution. The proof of Theorem \ref{thm-regen-bounded} exploits the fact that the underlying metric on $\xfancy$ is the Euclidean distance $d(\xvector, \xvector') = \norm{\xvector - \xvector'}$, which is also used to define the reconstruction loss $\lossrec$ (see Equation \ref{eq-lossrec}). The full proof can be found in Appendix \ref{sec-app-gen}.

The following theorem provides an upper bound of the distance between the input distribution and the VAE's generated distribution.
\begin{theorem}\label{thm-gen-bounded}
Under the definitions and assumptions of Theorem \ref{thm-rec-bounded}, we have that with probability at least $1-\delta$ over the random draw of $S\sampled \mu\otimesn$, the following holds for any posterior $\qphi{\zgivenx}$:
\begin{equation*}
\begin{split}
W_1(\mu, \pushf{\gthetaseul}{p(\zvector)}) \leq  \frac{1}{n} \somme{i=1}{n}{ \left\{ \expectseul{ \qphi{\zvector|\xvector_i}   }{\lossrec(\zvector, \xvector_i)}  \right\}    }  
  + 
\unsur{\lambda} \left( \somme{i=1}{n}{\kl{\qphi{\zvector|\xvector_i}}{p(\zvector)}  }  + \right. \\ 
  \left.
\log \unsur{\delta} + \frac{\lambda^2 \Delta^2}{8n}   \right) +
\frac{\ktheta}{n} \somme{i=1}{n}{\sqrt{ \norm{\muphi{\xvector_i}}^2 + \norm{\sigmaphi{\xvector_i} - \unvec}^2  }},
\end{split}
\end{equation*}
where $\Vec{1} \in \real^\dimz$ denotes the vector whose entries are all $1$.
\end{theorem}

The right-hand side of Theorem \ref{thm-gen-bounded} is equal to the right-hand side of Theorem \ref{thm-regen-bounded}, plus an additional term depending on the Wasserstein-2 distance $W_2(\qphi{\zgivenx_i}, p(\zvector))$, which is used in the proof because of its closed form for Gaussian distributions. 
Hence, the right-hand side of Theorem \ref{thm-gen-bounded} augments the VAE's optimization objective with $W_2(\qphi{\zgivenx_i}, p(\zvector))$, suggesting that a good generative performance may require the latent codes to be even closer to the prior. 
This is consistent with the findings of \citet{info-vae}, who showed that in order to improve generative performance, the latent codes need to be much closer to the prior, which may disrupt the balance between reconstruction loss and KL-loss.

\subsection{Regeneration and Generation Guarantees under the Manifold Assumption}
Similar to what we did in Section \ref{sec-recons-manif}, we assume that the data-generating distribution is a smooth transformation of the standard Gaussian distribution on $\real^\dimetoil$, where $\dimetoil$ is the intrinsic dimension of the dataset. This yields the following results.

\begin{theorem} \label{thm-regen-manif}
Under the definitions and assumptions of Theorem \ref{thm-rec-manif}, with probability at least $1 -\delta$ over the random draw of $S\sampled \mu\otimesn$, the following holds for any posterior $\qphi{\zgivenx}$:
\begin{equation*}
\begin{split}
W_1(\mu, \muhatn) \leq
 \frac{1}{n} \somme{i=1}{n}{ \left\{ \expectseul{ \qphi{\zvector|\xvector_i}   }{\lossrec(\zvector, \xvector_i)}  \right\}    }  
  + \unsur{\lambda} \left( 
  \somme{i=1}{n}{\kl{\qphi{\zvector|\xvector_i}}{p(\zvector)}  }  + 
\log \unsur{\delta} + \frac{\lambda^2 \ketoil^2}{2n} \right).
\end{split}
\end{equation*}
\end{theorem}

Note that the intrinsic and extrinsic dimensions do not explicitly appear in this inequality, although they may affect the reconstruction and KL loss.

We now present our last result, an upper bound on the Wasserstein distance between the input distribution and the VAE's generated distribution, under the manifold assumption.
\begin{theorem} \label{thm-gen-manif}
Under the definitions and assumptions of Theorem \ref{thm-rec-manif}, with probability at least $1 - \delta$ over the random draw of $S\sampled \mu\otimesn$, the following holds for any posterior $\qphi{\zgivenx}$ :
\begin{equation*}
\begin{split}
W_1(\mu, \pushf{\gthetaseul}{p(\zvector)}) \leq 
 \frac{1}{n} \somme{i=1}{n}{ \left\{ \expectseul{ \qphi{\zvector|\xvector_i}   }{\lossrec(\zvector, \xvector_i)}  \right\}    }  
  + \unsur{\lambda} \left( 
  \somme{i=1}{n}{\kl{\qphi{\zvector|\xvector_i}}{p(\zvector)}  }  + {} \right. \\
   \left.
\log \unsur{\delta} + \frac{\lambda^2 \ketoil^2}{2n} \right) + 
\frac{\ktheta}{n} \somme{i=1}{n}{\sqrt{ \norm{\muphi{\xvector_i}}^2 + \norm{\sigmaphi{\xvector_i} - \Vec{1}}^2  }},
\end{split}
\end{equation*}
where $\Vec{1} \in \real^\dimz$ denotes the vector whose entries are all $1$.
\end{theorem}

Theorem~\ref{thm-gen-bounded} and 
Theorem~\ref{thm-gen-manif} show that by minimizing the VAE's objective, one is also minimizing the Wasserstein distance between the input distribution and the VAE's generated distribution. 

From the upper bounds given by Theorems \ref{thm-regen-bounded} and \ref{thm-regen-manif}, one can deduce rates of convergence of $O(n^{-1/2})$ (when $\lambda \approx \sqrt{n}$) for the empirical regenerated distribution. Note that $\lambda \approx n$ leads to the much faster rate of $n^{-1}$, but then the bounds do not converge to the empirical risk, but to a larger positive number, dependent on the input distribution.
Similarly, Theorems~\ref{thm-gen-bounded} and \ref{thm-gen-manif} provide rates of convergence of $O(n^{-1/2})$ for the VAE's generated distribution.

\section{Discussion and Conclusion}\label{sec-discuss}

\paragraph{The different treatments of $\theta$ and $\phi$.} The bounds we've presented in this work hold for a given decoder $\theta$, but uniformly for all encoders. In practice, this means that the risk certificate has to be computed using samples different from the ones used to train the VAE. 
This is different from the usual PAC-Bayesian trick \citep[see also Remark~\ref{rem-prior-learning}]{pac-bayes-linear,pb-classif,perez-ortiz} of splitting the training set to learn the prior, then training the model on the whole training set, because the decoder and encoder are jointly optimized. Instead, one has to make sure that the model is only trained on samples distinct from the ones used to compute the bound. 
The same method would be necessary when computing the risk certificates given by the recent PAC-Bayesian bounds of \citet{beyond-usual} and \citet{online-pb}, since those bounds are not uniformly valid for any posterior.

\paragraph{The reconstruction loss.} In our bounds, the reconstruction loss is the $L_2$ norm (RMSE), instead of the squared $L_2$ norm (MSE). In practice, one can still optimize a VAE with the MSE (or any other reconstruction loss, e.g. the cross entropy loss), and then compute the bounds using the RMSE. However, if the reconstruction loss is not the RMSE, then the optima of the chosen optimization objective might differ from the ones minimizing the right-hand side of the bounds. Therefore, if the goal is to minimize the bounds, one should utilize the RMSE as the reconstruction loss.

\paragraph{Conclusion.} It is common, when applying PAC-Bayesian theory to new problems, to add additional stochasticity in order to account for the PAC-Bayesian distributions on the hypothesis class. For instance, \citet{pb-gen-models} added distributions on the parameters of a WGAN's generator, in order to perform a PAC-Bayesian analysis. However, because of the seamless integration of the PAC-Bayesian and VAE frameworks, such modification to the original problem has been avoided in this work. We matched the prior and posterior distributions on the VAE's latent space to the PAC-Bayesian prior and posterior, which allowed us to recover the VAE's optimization objective. We provide preliminary experiments on synthetic datasets in the supplementary material.

This work is a humble contribution to the theoretical understanding of VAEs. We developed novel PAC-Bayesian bounds suited to the analysis of VAEs and provided generalizations bounds for the VAE's reconstruction loss. In addition, we also derived upper bounds on the Wasserstein distance between the input distribution and the VAE's generative model's distribution. These bounds depend on the VAE's empirical optimization objective and the data-generating process. By integrating the VAE and PAC-Bayesian frameworks, we hope to establish PAC-Bayesian theory as a prime tool for the theoretical analysis of VAEs.

\section*{Acknowledgements}
This research is supported by the Canada CIFAR AI Chair Program, and the NSERC Discovery grant RGPIN-2020- 07223. F. Clerc is funded by IVADO through the DEEL Project CRDPJ 537462 18 and by a grant from NSERC.


\bibliographystyle{apalike}  
\bibliography{ref}  


\newpage
\title{Statistical Guarantees for Variational Autoencoders using PAC-Bayesian Theory: Supplementary Material}
\author{%
  Sokhna Diarra Mbacke \\
  Université Laval\\
  \texttt{sokhna-diarra.mbacke.1@ulaval.ca} \\
  \And
  Florence Clerc \\
  McGill University \\
  \texttt{florence.clerc@mail.mcgill.ca} \\
  \AND
  Pascal Germain \\
  Université Laval \\
  \texttt{pascal.germain@ift.ulaval.ca} \\
}
\date{}

\numberwithin{equation}{section}

\settitle

\appendix

\section{Preliminaries} \label{sec-app-basics}

\begin{definition}[Coupling]\label{def-coupling}
Let $p, q \in \mprob{\xfancy}$. A distribution $\gamma$ on $\xfancy\times \xfancy$ is a coupling \citep{villani}  of $p$ and $q$ if for every measurable set $B \subset \xfancy$, $\gamma(B \times \xfancy) = p(B)$ and $\gamma(\xfancy \times B) = q(B)$. In other words, a coupling of $p$ and $q$ is a distribution on $\xfancy\times \xfancy$ whose marginals are $p$ and $q$ respectively.

For example, the product measure $p \otimes q$ is a coupling of $p$ and $q$.
\end{definition}

\begin{definition}[Wasserstein distances]\label{def-app-wk}
Let $(\xfancy, d)$ be a Polish metric space and $p, q \in \mprob{\xfancy}$  Given a real number $k \geq 1$, the Wasserstein-$k$ distance $W_k$ is defined as
\[
W_k(p, q) = \left( \inf_{\pi \in \Gamma(p, q)}\integral{}{}{d(\xvector, \yvector)^k}{d\pi(\xvector, \yvector)}  \right)^{1/k},
\]
where $\Gamma(p, q)$ denotes the set of couplings of $p$ and $q$ (see Definition \ref{def-coupling} above). As stated in the main paper, $W_1$ is referred to as the Wasserstein distance.
\end{definition}

Given two Gaussian distributions $p=\normal{\mu_1, \Sigma_1}$ and $q=\normal{\mu_2, \Sigma_2}$ on $\real^\dimetoil$, the Wasserstein-2 distance has the following closed form \citep{w2-gaussian}:
\begin{equation}\label{eq-w2-gaussian}
W_2(p, q)^2 =  \norm{\mu_1 - \mu_2}^2 + \trace{\Sigma_1 + \Sigma_2 - 2\left( \Sigma_1^{1/2} \Sigma_2 \Sigma_1^{1/2} \right)^{1/2} }.
\end{equation}
This expression can be greatly simplified when the distributions have diagonal covariance matrices. Indeed, if $\Sigma_1 = \diag{\sigma_1^2}$ and $\Sigma_2 = \diag{\sigma_2^2}$ where $\sigma_1, \sigma_2 \in \real^\dimetoil$, then the product of the covariance matrices commutes $\Sigma_1 \Sigma_2 = \Sigma_2 \Sigma_1$ and we get
\[
\left( \Sigma_1^{1/2} \Sigma_2 \Sigma_1^{1/2} \right)^{1/2} = 
\Sigma_1^{1/2} \Sigma_2^{1/2},
\]
which, combined with the symmetry of covariance matrices and the definition of the Frobenius norm $\frobnorm{\cdot}$ \citep{cookbook}, implies 
\[
\trace{\Sigma_1 + \Sigma_2 - 2\left( \Sigma_1^{1/2} \Sigma_2 \Sigma_1^{1/2} \right)^{1/2} } = \frobnorm{\Sigma_1^{1/2} - \Sigma_2^{1/2}}^2 = \norm{\sigma_1 - \sigma_2}^2.
\]
Hence, if $p = \normal{\mu_1, \diag{\sigma_1^2}}$ and $q=\normal{\mu_2, \diag{\sigma_2^2}}$, then the Wasserstein-2 distance between $p$ and $q$ is
\begin{equation}\label{eq-w2-gaussian-diag}
W_2(p, q) = \norm{\mu_1 - \mu_2}^2 + \norm{\sigma_1 - \sigma_2}^2.
\end{equation}
We will use this equality to prove some of the results of Section \ref{sec-gen}.

The following change of measure theorem dates back to \cite{dv-paper} and has been used in the proof of many PAC-Bayesian theorems. A proof can be found in \citet[Corollary 4.15]{boucheron}.
\begin{proposition}[Donsker-Varadhan change of measure]\label{prop-dv}
Let $p, q$ be probability measures on a space $\hfancy$ such that $q \abscont p$, and let $g : \hfancy \rightarrow \real$ be a function such that $\expectseul{h\sampled p}{\expon{g(h)}} < \infty$. Then,
\[
\expectseul{h\sampled p}{\expon{g(h)}} \geq \expon{\expect{h\sampled q}{g(h)}  - \kl{q}{p}}.
\]
\end{proposition}
There are many different formulations of this proposition, we chose a formulation that facilitates readability of the proof of the following lemma.

\section{Proofs of the results in Section \ref{sec-pb-bound}} \label{sec-app-pb-bound}
We state and prove our first result. Note that the following lemma does not use Assumption \ref{assum-ipm}. Moreover, the main difference between the inequality of this lemma and the one of Theorem \ref{thm-after-ipm-assum} is the left-hand side. In Lemma \ref{lem-before-ipm-assum}, the expected loss for samples $\xvector\sampled \mu$  is computed w.r.t. distributions $q(h|\xvector_i)$ associated to the training samples. In contrast, in Theorem~\ref{thm-after-ipm-assum}, the expected loss for each $\xvector \sampled \mu$ is computed w.r.t. the distribution $q(h|\xvector)$ associated to $\xvector$ itself.
\begin{lemma}\label{lem-before-ipm-assum}
Let $\xfancy$ be the instance space, $\mu \in \mprob{\xfancy}$ the data-generating distribution, $\hfancy$ the hypothesis class, $\loss: \hfancy \times \xfancy \rightarrow \real$ the loss function, $p(h) \in \mprob{\hfancy}$ the prior distribution and $\delta \in (0, 1), \lambda >0$ real numbers. Then with probability at least $1-\delta$ over the random draw of the training set $S=\theset{\listen{\xvector}}\sampled \mu\otimesn$, the following holds for any conditional posterior $q(h|\xvector) \in \mprob{\hfancy}$:
\begin{equation}\label{eq-before-ipm-assum}
\begin{split}
 \frac{1}{n} \somme{i=1}{n}{ \left\{  \expectseul{h\sampled q(h|\xvector_i)}{\expectseul{ \xvector\sampled \mu }{\loss(h, \xvector)}}   \right\}    }  \leq
 \frac{1}{n} \somme{i=1}{n}{ \left\{ \expectseul{h \sampled q(h|\xvector_i)}{\loss(h, \xvector_i)}  \right\}    } +
   \frac{1}{\lambda} \left[ \somme{i=1}{n}{\kl{q(h|\xvector_i)}{p(h)}  }  
+   \right. \\  \left.
\log \unsur{\delta} + n \log  \expectseul{\xvector \sampled \mu}{} \expectseul{h\sampled p(h)}{  \exponbig{ \frac{\lambda}{n}  \left( \expect{\xvector'\sampled \mu}{\loss(h, \xvector')} -   \loss(h, \xvector)    \right)     }    }     \right].    
\end{split}
\end{equation}
\end{lemma}

\begin{proof}
First, we consider a set $H=\theset{\listen{h}} \sampled p(h)\otimesn$ iid sampled from $p(h)$. By applying Markov's inequality to the positive random variable $Y$, defined as
\[
Y \equdef \expectseul{H\sampled p(h)\otimesn}{\exponbig{ \frac{\lambda}{n} \somme{i=1}{n}{  \left\{ \expect{\xvector\sampled \mu}{\loss(h_i, \xvector)} -   \loss(h_i, \xvector_i)      \right\}  }    } },
\]
we obtain that with probability at least $1-\delta$ over the draw of $S\sampled \mu\otimesn$, $Y \leq \unsur{\delta}\expect{}{Y}$, meaning 
\begin{equation} \label{eq-after-markov}
\begin{split}
  &\hspace{-2.3cm}\expectseul{H\sampled p(h)\otimesn}{\exponbig{ \frac{\lambda}{n} \somme{i=1}{n}{  \left\{ \expect{\xvector\sampled \mu}{\loss(h_i, \xvector)} -   \loss(h_i, \xvector_i)      \right\}  }    }   }
\leq \\
&\unsur{\delta}\expectseul{S\sampled \mu\otimesn}{   \expectseul{H\sampled p(h)\otimesn}{\exponbig{ \frac{\lambda}{n} \somme{i=1}{n}{  \left\{ \expect{\xvector\sampled \mu}{\loss(h_i, \xvector)} -   \loss(h_i, \xvector_i)      \right\}  }    }   }      }.
\end{split}
\end{equation}
Let us focus on the left-hand side of \eqref{eq-after-markov}. We have
\begin{equation*}
\setlength{\jot}{10pt}
\begin{split}
    & \phantom{bbb} \expectseul{H\sampled p(h)\otimesn}{\exponbig{ \frac{\lambda}{n} \somme{i=1}{n}{  \left\{ \expect{\xvector\sampled \mu}{\loss(h_i, \xvector)} -   \loss(h_i, \xvector_i)      \right\}  }    }   } \\
     & =   \expectseul{H\sampled p(h)\otimesn}{ \produit{i=1}{n}{\exponbig{\frac{\lambda}{n}  \left( \expect{\xvector\sampled \mu}{\loss(h_i, \xvector)} -   \loss(h_i, \xvector_i)    \right)     }  }    }         \\
    & = \produit{i=1}{n}{\expectseul{h_i\sampled p(h)}{  \exponbig{ \frac{\lambda}{n}  \left( \expect{\xvector\sampled \mu}{\loss(h_i, \xvector)} -   \loss(h_i, \xvector_i)    \right)     }    }  }           \\
    & = \produit{i=1}{n}{\expectseul{h\sampled p(h)}{  \exponbig{ \frac{\lambda}{n}  \left( \expect{\xvector\sampled \mu}{\loss(h, \xvector)} -   \loss(h, \xvector_i)    \right)     }    }  }           \\
    & \geq  \produit{i=1}{n}{  \exponbig{ \expect{h\sampled q(h|\xvector_i)}{ \frac{\lambda}{n}  \left( \expect{\xvector\sampled \mu}{\loss(h, \xvector)} -   \loss(h, \xvector_i)    \right)   }  - \kl{q(h|\xvector_i)}{p(h)}}  }      ,
\end{split}
\end{equation*}
where the inequality uses the Donsker-Varadhan change of measure theorem (Proposition~\ref{prop-dv}). Applying the logarithm, we obtain
\begin{equation*}
\setlength{\jot}{10pt}
\begin{split}
    & \phantom{bbb} \log  \expectseul{H\sampled p(h)\otimesn}{\exponbig{ \frac{\lambda}{n} \somme{i=1}{n}{  \left\{ \expect{\xvector\sampled \mu}{\loss(h_i, \xvector)} -   \loss(h_i, \xvector_i)      \right\}  }    }   } \\
    & \geq \log \produit{i=1}{n}{  \exponbig{ \expect{h\sampled q(h|\xvector_i)}{ \frac{\lambda}{n}  \left( \expect{\xvector\sampled \mu}{\loss(h, \xvector)} -   \loss(h, \xvector_i)    \right)   }  - \kl{q(h|\xvector_i)}{p(h)} }  }  \\
    & = \somme{i=1}{n}{ \left( \expect{h\sampled q(h|\xvector_i)}{ \frac{\lambda}{n}  \left( \expect{\xvector\sampled \mu}{\loss(h, \xvector)} -   \loss(h, \xvector_i)    \right)   }  - \kl{q(h|\xvector_i)}{p(h)}   \right) }\\
    & = \frac{\lambda}{n} \somme{i=1}{n}{ \expect{h\sampled q(h|\xvector_i)}{ \expect{\xvector\sampled \mu}{\loss(h, \xvector)} -   \loss(h, \xvector_i)     } } - \somme{i=1}{n}{ \kl{q(h|\xvector_i)}{p(h)}  }.
\end{split}
\end{equation*}
This, combined with \eqref{eq-after-markov} yields 
\begin{equation}\label{eq-after-change-measure}
\begin{split}
\frac{\lambda}{n} \somme{i=1}{n}{ \expect{h\sampled q(h|\xvector_i)}{ \expect{\xvector\sampled \mu}{\loss(h, \xvector)} -   \loss(h, \xvector_i)     } } - \somme{i=1}{n}{ \kl{q(h|\xvector_i)}{p(h)}  } 
\leq  \\
\log \unsur{\delta}\expectseul{S\sampled \mu\otimesn}{   \expectseul{H\sampled p(h)\otimesn}{\exponbig{ \frac{\lambda}{n} \somme{i=1}{n}{  \left\{ \expect{\xvector\sampled \mu}{\loss(h_i, \xvector)} -   \loss(h_i, \xvector_i)      \right\}  }    }   }      }.
\end{split}
\end{equation}
It remains to show that the exponential moment on the right-hand side of Equation \eqref{eq-after-change-measure} can be modified by replacing the expectation w.r.t. $p(h)\otimesn$ with an expectation w.r.t. $p(h)$. Similar to what we did in the first part of the first derivation, we can use Fubini's theorem to obtain
\begin{equation*}
\setlength{\jot}{10pt}
\begin{split}
    & \phantom{bbb} \expectseul{S\sampled \mu\otimesn}{}  \expectseul{H\sampled p(h)\otimesn}{\exponbig{ \frac{\lambda}{n} \somme{i=1}{n}{  \left\{ \expect{\xvector\sampled \mu}{\loss(h_i, \xvector)} -   \loss(h_i, \xvector_i)      \right\}  }    }   } \\
    & = \expectseul{S\sampled \mu\otimesn}{}  \produit{i=1}{n}{\expectseul{h\sampled p(h)}{  \exponbig{ \frac{\lambda}{n}  \left( \expect{\xvector\sampled \mu}{\loss(h, \xvector)} -   \loss(h, \xvector_i)    \right)     }    }  }           \\
    & =  \produit{i=1}{n}{ \expectseul{\xvector_i \sampled \mu}{} \expectseul{h\sampled p(h)}{  \exponbig{ \frac{\lambda}{n}  \left( \expect{\xvector\sampled \mu}{\loss(h, \xvector)} -   \loss(h, \xvector_i)    \right)     }    }  }        \\
    & =  \produit{i=1}{n}{ \expectseul{\xvector \sampled \mu}{} \expectseul{h\sampled p(h)}{  \exponbig{ \frac{\lambda}{n}  \left( \expect{\xvector'\sampled \mu}{\loss(h, \xvector')} -   \loss(h, \xvector)    \right)     }    }  }   \\
    & = \left( \expectseul{\xvector \sampled \mu}{} \expectseul{h\sampled p(h)}{  \exponbig{ \frac{\lambda}{n}  \left( \expect{\xvector'\sampled \mu}{\loss(h, \xvector')} -   \loss(h, \xvector)    \right)     }    }    \right)^n
    .
\end{split}
\end{equation*}
Hence, 
\begin{equation*}
\begin{split}
\hspace{-3cm}\log \expectseul{S\sampled \mu\otimesn}{}  \expectseul{H\sampled p(h)\otimesn}{\exponbig{ \frac{\lambda}{n} \somme{i=1}{n}{  \left\{ \expect{\xvector\sampled \mu}{\loss(h_i, \xvector)} -   \loss(h_i, \xvector_i)      \right\}  }    }   }  =
\\
\hspace{2cm} n \log \expectseul{\xvector \sampled \mu}{} \expectseul{h\sampled p(h)}{  \exponbig{ \frac{\lambda}{n}  \left( \expect{\xvector'\sampled \mu}{\loss(h, \xvector')} -   \loss(h, \xvector)    \right)     }    } .  
\end{split}
\end{equation*}


Combining this equation with Equation~\eqref{eq-after-change-measure} yields the theorem.
\end{proof}
The reader familiar with PAC-Bayes bounds may notice that the proof of Lemma \ref{lem-before-ipm-assum} is similar to the usual derivation of PAC-Bayesian bounds, with a key difference. We start with an iid set of $n$ hypotheses sampled from the prior, which allows us to apply the change of measure theorem to $n$ posteriors $q(h|\xvector_1), \dots, q(h|\xvector_n)$. Then, we show that the exponential moment obtained with $n$ hypotheses instead of one is equal to the exponential moment obtained with one hypothesis.

\subsection{Proof of Theorem \ref{thm-after-ipm-assum}}
The first summand on the left-hand side of Lemma \ref{lem-before-ipm-assum} is the risk on samples $\xvector\sampled \mu$, when the hypotheses are uniformly sampled from $q(h|\xvector_i), 1\leq i \leq n$. In order to replace $q(h|\xvector_i)$ by $q(h|\xvector)$ in that term and derive Theorem \ref{thm-after-ipm-assum}, we utilize Assumption \ref{assum-ipm}.

First, recall that Theorem \ref{thm-after-ipm-assum} states that under the assumptions of Lemma \ref{lem-before-ipm-assum}, if Assumption \ref{assum-ipm} holds with a constant $K>0$, then the following inequality holds with probability at least $1-\delta$:
\begin{equation}\label{eq-app-after-ipm-assum}
\begin{split}
\expectseul{\xvector\sampled \mu}{\expectseul{h\sampled q(h|\xvector)}{\loss(h, \xvector)}} - \unsur{n}  \somme{i=1}{n}{ \expectseul{h\sampled q(h|\xvector_i)}{\loss(h, \xvector_i)}  }  
  \leq \unsur{\lambda} \left[ 
  \somme{i=1}{n}{\kl{q(h|\xvector_i)}{p(h)}  }  
+ \frac{\lambda K}{n} \somme{i=1}{n}{\expect{\xvector\sampled \mu}{d(\xvector, \xvector_i)}} + \right.
\\  \left.
\log \unsur{\delta} + n \log  \expectseul{\xvector \sampled \mu}{} \expectseul{h\sampled p(h)}{  \expon{ \frac{\lambda}{n}  \left( \expect{\xvector'\sampled \mu}{\loss(h, \xvector')} -   \loss(h, \xvector)    \right)     }    }   \right].    
\end{split}
\end{equation}

\begin{proof}[Proof of Theorem \ref{thm-after-ipm-assum}]
Using the definition of an IPM and Assumption \ref{assum-ipm}, for any $\xvector_i \in S, \xvector \in \xfancy$, we have
\[
\expectseul{h\sampled q(h|\xvector)}{\loss(h, \xvector)} - \expectseul{h\sampled q(h|\xvector_i)}{\loss(h, \xvector)}  \leq \ipm{\newf } (q(h|\xvector), q(h|\xvector_i)) \leq K d(\xvector, \xvector_i).
\]
Combined with Fubini's theorem, we obtain
\[
\somme{i=1}{n}{  \expectseul{h\sampled q(h|\xvector_i)}{\expectseul{ \xvector\sampled \mu }{\loss(h, \xvector)}}      }  = \somme{i=1}{n}{  \expect{\xvector\sampled \mu}{\expectseul{ h\sampled q(h|\xvector_i) }{\loss(h, \xvector)}} }  \geq
\somme{i=1}{n}{\expect{\xvector\sampled \mu}{\expectseul{h\sampled q(h|\xvector)}{\loss(h, \xvector)} - Kd(\xvector, \xvector_i)   }} .
\]
Combining this with Lemma \ref{lem-before-ipm-assum}, yields Theorem \ref{thm-after-ipm-assum}.
\end{proof}


\section{Proofs of the results in Section \ref{sec-recons}} \label{sec-app-recons}
\subsection{Proof of Proposition \ref{prop-assum-ipm}}

First, we recall the statement of Proposition \ref{prop-assum-ipm}.
\begin{proposition}[Restatement of Proposition \ref{prop-assum-ipm}]
If there exists positive real numbers $\kphi$ and $\ktheta$ such that the encoder and decoder are respectively $\kphi$-Lipschitz and $\ktheta$-Lipschitz continuous, then 
\begin{equation}\label{eq-app-ipm-df}
\ipm{\newf}\left( \qphi{\zgivenx_1}, \qphi{\zgivenx_2}  \right) \leq \kphi \ktheta \norme{\xvector_1 - \xvector_2},
\end{equation}
and 
\begin{equation}\label{eq-app-ipm-loss}
\loss(\cdot, \xvector) \in \newf, \quad \text{for any } \xvector\in \xfancy.
\end{equation}
where $\newf = \lipschitz{\ktheta}(\zfancy, \real)$ is the set of real-valued $\ktheta$-Lipschitz continuous functions defined on $\zfancy$.
\end{proposition}

\begin{proof} \mbox{}
\begin{enumerate}
\item Let us prove \eqref{eq-app-ipm-df}. First, since $\qphi{\zgivenx_i} = \normal{\muphi{\xvector_i}, \diag{\sigmaphisq{\xvector_i}}}$, by \eqref{eq-w2-gaussian-diag}, the Wasserstein-2 distance $W_2(\qphi{\zgivenx_1}, \qphi{\zgivenx_2})$ has the following closed form:
\[
W_2(\qphi{\zgivenx_1}, \qphi{\zgivenx_2})^2 = \norme{\muphi{\xvector_1} -\muphi{\xvector_2}}^2  + \norme{\sigmaphi{\xvector_1} - \sigmaphi{\xvector_2}}^2,
\]
which, combined with the definition $\encfunc{\xvector} = \begin{bmatrix}
   \muphi{\xvector} \\
   \sigmaphi{\xvector}
 \end{bmatrix}$, yields 
\[
\norme{\encfunc{\xvector_1} - \encfunc{\xvector_2}}^2 =  W_2(\qphi{\zgivenx_1}, \qphi{\zgivenx_2})^2 .
\]
Since $\encfuncseul$ is $\kphi$-Lipschitz continuous, we have $\norme{\encfunc{\xvector_1} - \encfunc{\xvector_2}} \leq \kphi \norme{\xvector_1 - \xvector_2}$, and
\begin{equation}\label{eq-app-w2}
W_2(\qphi{\zgivenx_1}, \qphi{\zgivenx_2}) \leq \kphi \norme{\xvector_1 - \xvector_2}.
\end{equation}
On the other hand, the definition $\newf = \lipschitz{\ktheta}(\zfancy, \real) $ and the Kantorovich duality imply
\[
\ipm{\newf}( \qphi{\zgivenx_1}, \qphi{\zgivenx_2} ) = \ktheta W_1(\qphi{\zgivenx_1}, \qphi{\zgivenx_2}).
\]
Since $W_1 \leq W_2$, this equation, combined with \eqref{eq-app-w2} yields 
\[
\ipm{\newf}( \qphi{\zgivenx_1}, \qphi{\zgivenx_2} ) \leq \ktheta \kphi \norme{\xvector_1 - \xvector_2}.
\]

\item Now, we shall prove \eqref{eq-app-ipm-loss}, meaning, we show that $\loss(\cdot, \xvector) \in \lipschitz{\ktheta}(\zfancy, \real)$ . Let $\xvector \in \xfancy$ and $\zvector_1, \zvector_2 \in \zfancy$. We have
\begin{equation*}
\setlength{\jot}{10pt}
\begin{split}
\loss(\zvector_1, \xvector) - \loss(\zvector_2, \xvector) & = \norm{\xvector - \gtheta{\zvector_1}} - \norm{\xvector - \gtheta{\zvector_2}}           \\
    & = \norm{\xvector - \gtheta{\zvector_1} +\gtheta{\zvector_2} - \gtheta{\zvector_2}} - \norm{\xvector - \gtheta{\zvector_2}}             \\
    & \leq \norm{\xvector - \gtheta{\zvector_2}} + \norm{\gtheta{\zvector_2} - \gtheta{\zvector_1}} - \norm{\xvector - \gtheta{\zvector_2}}          \\
    & =   \norm{\gtheta{\zvector_2} - \gtheta{\zvector_1}}         \\
    & \leq \ktheta \norm{\zvector_1 - \zvector_2},          
\end{split}
\end{equation*}
where the first inequality uses the triangle inequality and the second uses the Lipschitz assumption on $\gthetaseul$.

\end{enumerate}
\end{proof}

\subsection{Proof of Theorem \ref{thm-rec-bounded}}
\label{sec-proof-thm-rec-bounded}


\begin{proof}[Proof of Theorem \ref{thm-rec-bounded}]
In order to prove Theorem \ref{thm-rec-bounded}, we need to upper bound the average distance and the exponential moment of Theorem \ref{thm-gen-vae}, under the finite diameter assumption:
\begin{equation}\label{eq-app-diam}
\sup_{\xvector, \xvector' \in \xfancy} d(\xvector, \xvector') = \Delta < \infty.
\end{equation}
More precisely, we need to prove the two following inequalities.
\begin{equation} \label{eq-app-avg-bounded}
\somme{i=1}{n}{\expectseul{\xvector \sampled \mu}{d(\xvector, \xvector_i)}} \leq 
n\Delta
\end{equation}
and
\begin{equation} \label{eq-app-expon-bounded}
n \log \expectseul{\zvector \sampled p(\zvector)}{\expectseul{\xvector \sampled \mu}{\exponbig{\frac{\lambda}{n} \left( \expectseul{\xvector' \sampled \mu}{\lossrec(\zvector, \xvector')} - \lossrec(\zvector, \xvector) \right) }}} \leq 
\frac{\lambda^2 \Delta^2}{8n}.
\end{equation}


First, \eqref{eq-app-avg-bounded} is a direct consequence of the definition of the diameter $\Delta$.


Now, let us prove \eqref{eq-app-expon-bounded}. Let $\zvector \in \zfancy$. Since $\lossrec(\zvector, \xvector) = \norm{\xvector - \gtheta{\zvector}} = d(\xvector, \gtheta{\zvector})$ is the distance between $\xvector$ and $\gtheta{\zvector}$, the definition of $\Delta$ implies $\lossrec(\zvector, \xvector) \in [0, \Delta]$, for any $\xvector \in \xfancy$. Hence, applying Hoeffding's lemma on the random variables $\loss_i = \lossrec(\zvector, \xvector_i) \in [0, \Delta]$, we obtain
\[ 
\expectseul{\xvector \sampled \mu}{  \exponbig{\frac{\lambda}{n} \left( \expect{\xvector'\sampled \mu}{\lossrec(\zvector, \xvector')} - \lossrec(\zvector, \xvector)    \right)}      } \leq \exponbig{\frac{\lambda^2 \Delta^2}{8n^2}}.
\]
Which leads to
\[
n \log \expectseul{\zvector \sampled p(\zvector)}{\expectseul{\xvector \sampled \mu}{\exponbig{\frac{\lambda}{n} \left( \expectseul{\xvector' \sampled \mu}{\lossrec(\zvector, \xvector')} - \lossrec(\zvector, \xvector) \right) }}} \leq 
n \log \expectseul{\zvector\sampled p(\zvector)}{
\exponbig{\frac{\lambda^2 \Delta^2}{8n^2}}} = 
\frac{\lambda^2 \Delta^2}{8n}.
\]


\end{proof}

\subsection{Proof of Theorem \ref{thm-rec-manif}}\label{sec-app-rec-manif}
We need to bound the average distance and the exponential moment of Theorem \ref{thm-gen-vae}, under the assumption $\mu = \pushf{\getoilseul}{\petoil}$, with $\petoil = \normal{\zeromatrix, \idmatrix}$ is the standard Gaussian distribution on $\real^\dimetoil$, and $\getoilseul \in \lipschitz{\ketoil}(\real^\dimetoil, \xfancy)$.

\begin{lemma}
\label{lemma:exp-moment-thm-rec-manif}
Under the hypotheses of Theorem \ref{thm-rec-manif}, the following inequality holds:
\begin{equation} \label{eq-app-expon-manif}
n \log \expectseul{\zvector \sampled p(\zvector)}{\expectseul{\xvector \sampled \mu}{\exponbig{\frac{\lambda}{n} \left( \expectseul{\xvector' \sampled \mu}{\lossrec(\zvector, \xvector')} - \lossrec(\zvector, \xvector) \right) }}} \leq 
\frac{\lambda^2 \ketoil^2}{2n}.
\end{equation}

\end{lemma}

\begin{proof}
Let us show that
\[
n \log \expectseul{\zvector \sampled p(\zvector)}{\expectseul{\xvector \sampled \mu}{\exponbig{\frac{\lambda}{n} \left( \expectseul{\xvector' \sampled \mu}{\lossrec(\zvector, \xvector')} - \lossrec(\zvector, \xvector) \right) }}} \leq 
\frac{\lambda^2 \ketoil^2}{2n}.
\]
Since $\mu = \pushf{\getoilseul}{\petoil}$, where $\petoil$ is the standard Gaussian distribution on $\real^\dimetoil$ and $\getoilseul$ is $\ketoil$-Lipschitz continuous, the definition of the loss function $\lossrec$ implies
\begin{equation*}
\setlength{\jot}{10pt}
\begin{split}
 &  \expectseul{\xvector \sampled \mu}{\exponbig{\frac{\lambda}{n} \left( \expectseul{\xvector' \sampled \mu}{\lossrec(\zvector, \xvector')} - \lossrec(\zvector, \xvector) \right) }}   \\
    & = \expectseul{\xvector \sampled \mu}{\exponbig{\frac{\lambda}{n} \left( \expectseul{\xvector' \sampled \mu}{\norm{ \xvector' - \gtheta{\zvector}}} - \norm{\xvector - \gtheta{\zvector}} \right) }} .     \\
    & = \expectseul{\zetoil \sampled \petoil}{\exponbig{\frac{\lambda}{n} \left( \expectseul{\zetoil' \sampled \petoil}{\norm{ \getoil{\zetoil'} - \gtheta{\zvector}}} - \norm{ \getoil{\zetoil} - \gtheta{\zvector}} \right) }}         \\
    & \leqstar  \exponbig{\frac{\lambda^2 \ketoil^2}{2n^2}   }     . 
    \end{split}
\end{equation*}
This derivation implies
\[
n \log \expectseul{\zvector \sampled p(\zvector)}{\expectseul{\xvector \sampled \mu}{\exponbig{\frac{\lambda}{n} \left( \expectseul{\xvector' \sampled \mu}{\lossrec(\zvector, \xvector')} - \lossrec(\zvector, \xvector) \right) }}} \leq 
\frac{\lambda^2 \ketoil^2}{2n}.
\]

We still need to justify $\leqstar$. Define for any arbitrary $\alpha \in \xfancy$ the function $f : \real^\dimetoil \rightarrow \real$ as:
\[
f(\zetoil) = \norm{\getoil{\zetoil} - \alpha}.
\]
Since $g^* \in \lipschitz{\ketoil}(\real^\dimetoil, \xfancy)$, the function $f$
is $\ketoil$-Lipschitz. Indeed, for any $\zetoil_1, \zetoil_2 \in \real^\dimetoil$,
\begin{equation*}
\setlength{\jot}{10pt}
\begin{split}
f(\zetoil_1) - f(\zetoil_2) & =   \norm{\getoil{\zetoil_1} - \alpha} - \norm{\getoil{\zetoil_2} - \alpha}           \\
    & =   \norm{\getoil{\zetoil_1} - \alpha + \getoil{\zetoil_2} - \getoil{\zetoil_2}} - \norm{\getoil{\zetoil_2} - \alpha}           \\
    & \leq  \norm{\getoil{\zetoil_1} - \getoil{\zetoil_2}} + \norm{\getoil{\zetoil_2} - \alpha} - \norm{\getoil{\zetoil_2} - \alpha}             \\
    & =  \norm{\getoil{\zetoil_1} - \getoil{\zetoil_2}}         \\
    &  \leq  \ketoil \norm{\zetoil_1 - \zetoil_2}          \\
\end{split}
\end{equation*}
Moreover, it is known (see Theorem 5.5 of \citet{boucheron}) that if $f$ is a $\ketoil$-Lipschitz function of a standard normal random variable $\zvector$, then 
\[
\expectseul{}{ \expon{\lambda(\expectseul{}{[f(\zvector)]} - f(\zvector)   )}    } \leq \expon{\frac{\lambda^2 \ketoil^2}{2}}.
\]
Hence,
\[
\expect{\zetoil_i \sampled \petoil}{ \exponbig{\frac{\lambda}{n}  \left(  \expect{\zetoil' \sampled \petoil}{\norm{\getoil{\zetoil'} - \gtheta{\zvector}  }} - \norm{\getoil{\zetoil_i} - \gtheta{\zvector}}   \right)      }  }  \leq \exponbig{\frac{\lambda^2 \ketoil^2}{2n^2}},
\]
which proves $\leqstar$ and concludes this proof. 
\end{proof}


\begin{lemma}
\label{lemma:avg-dist-thm-rec-manif}
Under the hypotheses of Theorem \ref{thm-rec-manif}, with probability at least $1 -\frac{n\dimetoil}{2}\expon{\frac{-a^2}{2}}$ over the random draw of $S$, 
\begin{equation}
\somme{i=1}{n}{\expectseul{\xvector \sampled \mu}{d(\xvector, \xvector_i)}} \leq 
n \ketoil \sqrt{(1 + a^2)\dimetoil}
\end{equation}
\end{lemma}

\begin{proof}
First, since the training set $S = \theset{\listen{\xvector}} \iidsampled \mu$, for each $1 \leq i \leq n$, there exists $\zetoil_i \sampled \petoil$ such that $\xvector_i  = \getoil{\zetoil_i}$. Let $a>0$ be a positive real number. By definition of $\petoil$, we have
\[
\prob{}{\forall i, \zetoil_i \in [-a, a]^\dimetoil} = \left(\erf{\frac{a}{\sqrt{2}}} \right)^{n \dimetoil},
\]
where $\erf{\cdot}$ denotes the error function. Since the error function verifies (see \cite{chu1955bounds})
\[
\erf{\frac{a}{\sqrt{2}}}  \geq \sqrt{1-\expon{\frac{-a^2}{2}}},
\]
we can use Bernoulli's inequality (see Section 2.4 of \citet{analytic-ineq}) to obtain
\begin{equation} \label{eq-app-confidence}
\prob{}{\forall i, \zetoil_i \in [-a, a]^\dimetoil} \geq \left( 1-\expon{\frac{-a^2}{2}}  \right)^{n\dimetoil/2} \geq 1 -\frac{n\dimetoil}{2}\expon{\frac{-a^2}{2}}.
\end{equation}

Now we assume $\zetoil_i \in [-a, a]^\dimetoil$ for all $1\leq i \leq n$ and we shall prove the desired inequality:
\begin{equation} \label{eq-app-avg-manif}
\somme{i=1}{n}{\expectseul{\xvector \sampled \mu}{d(\xvector, \xvector_i)}} \leq 
n \ketoil\sqrt{(1 + a^2)\dimetoil}
\end{equation}

Let us prove \eqref{eq-app-avg-manif}. We have
\begin{equation}\label{eq-app-lipketoil}
\expectseul{\xvector \sampled \mu}{d(\xvector, \xvector_i)} = \expectseul{\xvector \sampled \mu}{\norm{\xvector- \xvector_i}} = \expectseul{\zetoil \sampled \petoil}{\norm{\getoil{\zetoil}- \getoil{\zetoil_i}}} \leq \ketoil \expectseul{\zetoil \sampled \petoil}{\norm{\zetoil- \zetoil_i}},
\end{equation}
where the inequality follows from the assumption $\getoilseul \in \lipschitz{\ketoil}(\real^\dimetoil, \xfancy)$. Using Holder's inequality, the fact that $\norm{\zetoil - \zetoil_i}^2$ is a non-central $\chi^2$ random variable with $\dimetoil$ degrees of freedom and non-centrality coefficient $\norm{\zetoil_i}^2$, and the assumption $\zetoil_i \in [-a, a]^\dimetoil$, we obtain
\[
\expectseul{\zetoil \sampled \petoil}{\norm{\zetoil- \zetoil_i}} \leq \left( \expectseul{\zetoil \sampled \petoil}{\norm{\zetoil- \zetoil_i}^2}  \right)^{1/2}
= \left( \dimetoil + \norm{\zetoil_i}^2  \right)^{1/2} \leq \left( \dimetoil + a^2\dimetoil  \right)^{1/2}.
\]
 Hence, 
\[
\expectseul{\xvector \sampled \mu}{\norm{\xvector - \xvector_i}} \leq \ketoil \sqrt{(1 + a^2)\dimetoil}
\]
which proves \eqref{eq-app-avg-manif}.
\end{proof}

\begin{proof}[Proof of Theorem \ref{thm-rec-manif}]
Lemmas \ref{lemma:exp-moment-thm-rec-manif} and \ref{lemma:avg-dist-thm-rec-manif} applied to the result from Theorem \ref{thm-gen-vae} provide us with the inequality of Theorem \ref{thm-rec-manif}. Finally, the confidence of $1-\delta -\frac{n\dimetoil}{2}\expon{\frac{-a^2}{2}} $ is obtained by using the union bound: the inequality in Theorem \ref{thm-gen-vae} holds with probability at least $1- \delta$, whereas the inequality appearing in Lemma \ref{lemma:avg-dist-thm-rec-manif} holds with probability at least $1 -\frac{n\dimetoil}{2}\expon{\frac{-a^2}{2}}$.
\end{proof}

In the following proposition, we provide an alternate version of Theorem \ref{thm-rec-manif}, where the distribution $\petoil$ is the uniform distribution\footnote{Note that the result holds for any distribution on $[0, 1]^\dimetoil$, not just the uniform distribution.} on $[0, 1]^\dimetoil$, instead of the standard Gaussian distribution on $\real^\dimetoil$.
\begin{proposition}\label{prop-rec-manif-uniform}
Let $\xfancy$ be the instance space, $\zfancy$ the latent space, $p(\zvector)  \in \mprob{\zfancy}$ the prior distribution, $\theta$ the parameters of the decoder, $\delta \in (0, 1), \lambda >0, a>0$ be real numbers. Assume the data-generating distribution $\mu = \pushf{\getoilseul}{\petoil}$, where $\petoil = \ufancy([0, 1]^\dimetoil)$ is the uniform distribution on $[0, 1]^\dimetoil$ and $\getoilseul \in \lipschitz{\ketoil}(\real^\dimetoil, \xfancy)$ is $\ketoil$-Lipschitz continuous. With probability at least $1 - \delta $ over the random draw of $S$, the following holds for any posterior $\qphi{\zgivenx}$: 
\begin{equation*}
\begin{split}
 \expectseul{\xvector\sampled \mu }{\expectseul{\qphi{\zvector|\xvector}  }{\lossrec(\zvector, \xvector)}}      -
 \frac{1}{n} \somme{i=1}{n}{ \left\{ \expectseul{ \qphi{\zvector|\xvector_i}   }{\lossrec(\zvector, \xvector_i)}  \right\}    }  
  \leq \unsur{\lambda} \left( 
  \somme{i=1}{n}{\kl{\qphi{\zvector|\xvector_i}}{p(\zvector)}  }  + \right. \\
   \lambda \kphi \ktheta \ketoil \sqrt{\dimetoil}  +  \left.
\log \unsur{\delta} + \frac{\lambda^2 \ketoil^2}{2n} \right).
\end{split}
\end{equation*}

\end{proposition}

\begin{proof}
Let $\theset{\listen{\zetoil}} \subseteq [0, 1]^\dimetoil$ be such that for all $1\leq i \leq n$, $\xvector_i = \getoil{\zetoil_i}$. Since the diameter of $[0, 1]^\dimetoil$ is $\sqrt{\dimetoil}$, using the assumptions on $\mu$ and $\getoilseul$, we obtain
\[
\somme{i=1}{n}{\expectseul{\xvector \sampled \mu}{d(\xvector, \xvector_i)}} = 
\somme{i=1}{n}{\expectseul{\zetoil \sampled \petoil}{d(\getoil{\zetoil}, \getoil{\zetoil_i})  }} \leq
\ketoil \somme{i=1}{n}{\expectseul{\zetoil \sampled \petoil}{\norm{\zetoil- \zetoil_i}  }} \leq n \ketoil \sqrt{\dimetoil}.
\]
Applying the inequality above to Theorem \ref{thm-gen-vae} yields the desired result.
\end{proof}
Note that unlike Theorem \ref{thm-rec-manif}, the confidence $1-\delta$ of Theorem \ref{thm-gen-vae} is not lowered in Proposition \ref{prop-rec-manif-uniform}.


\section{Proofs of the results in Section \ref{sec-gen}} \label{sec-app-gen}
To simplify the proofs of the theorems of Section~\ref{sec-gen}, we start by proving Lemmas \ref{lem-mu-muhatn} and \ref{lem-muhatn-gpz} below.

First, recall the definition of $\muhatn$:
\[
\muhatn = \unsur{n} \somme{i=1}{n}{\pushf{\gthetaseul}{\qphi{\zgivenx_i}}}   .
\]
The triangle inequality implies
\begin{equation}\label{eq-app-triangle-mu}
W_1(\mu, \pushf{\gthetaseul}{p(\zvector)}) \leq W_1(\mu, \muhatn) + W_1(\muhatn, \pushf{\gthetaseul}{p(\zvector)}).
\end{equation}

Let us state and prove the first lemma of this section.
\begin{lemma} \label{lem-mu-muhatn}
The following inequality holds with probability at least $1-\delta$ over the random draw of $S\sampled \mu\otimesn$:
\begin{equation*}
\begin{split}
\lambda W_1(\mu, \muhatn) \leq \frac{\lambda}{n} \somme{i=1}{n}{ \left( \expectseul{\zvector \sampled q(\zvector|\xvector_i)}{\lossrec(\zvector, \xvector_i)}  \right)    } 
  + \somme{i=1}{n}{\kl{q(\zvector|\xvector_i)}{p(\zvector)}  }  
+  \\
\log \unsur{\delta} + \log \expectseul{S\sampled \mu\otimesn}{   \expectseul{\zvector\sampled p(\zvector)}{\expon{ \lambda  \left( \expect{\xvector\sampled \mu}{\lossrec(\zvector, \xvector)} -  \unsur{n} \somme{i=1}{n}{ \lossrec(\zvector, \xvector_i)}      \right)  }    }        }.   
\end{split}
\end{equation*}
\end{lemma}

\begin{proof}
Recall the expression for the Wasserstein distance based on couplings:
\[ W_1(\mu, \muhatn) =    \inf_{\pi \in \Gamma(\mu, \muhatn)}  \integral{\xfancy\times \xfancy}{ }{ \norm{\xvector -\yvector} }{d\pi(\xvector, \yvector)}\]
In particular, $W_1 (\mu, \muhatn)$ is less than the right-hand side obtained by the product coupling which can be rewritten, using Fubini's theorem, as:
\begin{align*}
W_1(\mu, \muhatn)
    & \leq   \integral{\xfancy\times \xfancy}{ }{ \norm{\xvector -\yvector} }{  d\mu(\xvector) d\muhatn(\yvector)    }         \\
    & = \expectseul{\yvector \sampled \muhatn}{\, \expectseul{\xvector\sampled \mu}{ \norm{\xvector-\yvector}     }}        .
\end{align*}

Using the derivation above and the definition of $\muhatn$, we obtain
\begin{align*}
W_1(\mu, \muhatn) \leq
 \expectseul{\yvector\sampled \muhatn}{\, \expectseul{\xvector \sampled \mu}{ \norme{\xvector-\yvector}     }}  
 & = 
\unsur{n} \somme{i=1}{n}{\left( \expectseul{\zvector\sampled \qphi{\zgivenx_i}}{\expectseul{\xvector \sampled \mu}{ \norme{\xvector - \gtheta{\zvector}}    }  } \right) }  \\
& = 
 \unsur{n} \somme{i=1}{n}{\left( \expectseul{\zvector\sampled \qphi{\zgivenx_i}}{\expectseul{\xvector \sampled \mu}{ \lossrec(\zvector, \xvector)    }  } \right) } .
\end{align*}
We can upper bound this expression using Lemma \ref{lem-before-ipm-assum} with $\hfancy = \zfancy$ and $\loss = \lossrec$. We get that  with probability at least $1-\delta$ over the random draw of $S\sampled \mu\otimesn$:
\begin{equation*}\label{eq-use-lemma-before-ipm}
\begin{split}
 \frac{\lambda}{n} \somme{i=1}{n}{ \left(  \expectseul{\zvector \sampled q(\zvector|\xvector_i)}{\expectseul{ \xvector\sampled \mu }{\lossrec(\zvector, \xvector)}}   \right)    }  \leq
 \frac{\lambda}{n} \somme{i=1}{n}{ \left( \expectseul{\zvector \sampled q(\zvector|\xvector_i)}{\lossrec(\zvector, \xvector_i)}  \right)    } 
  + \somme{i=1}{n}{\kl{q(\zvector|\xvector_i)}{p(\zvector)}  }  
+  \\
\log \unsur{\delta} + \log \expectseul{S\sampled \mu\otimesn}{   \expectseul{\zvector\sampled p(\zvector)}{\expon{ \lambda  \left( \expect{\xvector\sampled \mu}{\lossrec(\zvector, \xvector)} -  \unsur{n} \somme{i=1}{n}{ \lossrec(\zvector, \xvector_i)}      \right)  }    }        }.    
\end{split}
\end{equation*}
\end{proof}
Therefore, using the upper bounds on the exponential moment from Section~\ref{sec-recons}, we can prove Theorems \ref{thm-regen-bounded} and \ref{thm-regen-manif} in the following sections.

Next, we prove the following lemma.
\begin{lemma}\label{lem-muhatn-gpz}
The following inequality holds.
\[
W_1(\muhatn, \pushf{\gthetaseul}{p(\zvector)}) \leq \frac{\ktheta}{n} \somme{i=1}{n}{ \sqrt{ \norme{\muphi{\xvector_i}}^2 + \norme{\sigmaphi{\xvector_i} - \unvec}^2} } ,
\]
where $\unvec \in \real^\dimz$ denotes the vector whose entries are all $1$.
\end{lemma}

\begin{proof}
Defining the mixture of measures
\[
\qhatn(\zvector) = \unsur{n}\somme{i=1}{n}{\qphi{\zgivenx_i}},
\]
the definition of $\muhatn$ and the definition of a pushforward measures yield
\[
\muhatn = \unsur{n} \somme{i=1}{n}{\pushf{\gthetaseul}{\qphi{\zgivenx_i}}}  = \pushf{\gthetaseul}{\qhatn(\zvector)}.
\]
Using the dual formulation of the Wasserstein distance, we have
\begin{equation*}
\setlength{\jot}{10pt}
\begin{split}
W_1(\muhatn, \pushf{g_\theta }{p(\zvector)})    & = W_1\left(   \pushf{g_\theta}{\qhatn(\zvector) } , \pushf{g_\theta}{p(\zvector)}  \right)          \\
    & =   \sup_{f \in \lipschitz{1}(\xfancy, \real)}\left[  \integral{\zfancy}{}{f\circ g_\theta (\zvector)}{d\qhatn(\zvector)}  -  \integral{\zfancy}{}{f\circ g_\theta (\zvector)}{dp(\zvector)}  \right]           \\
    & =  \sup_{g \in \gfancy_\theta}\left[  \integral{\zfancy}{}{g (\zvector)}{d\qhatn(\zvector)}  -  \integral{\zfancy}{}{g (\zvector)}{dp(\zvector)}  \right]          \\
    & \leq  \sup_{g \in \lipschitz{\ktheta}(\zfancy, \real)}\left[  \integral{\zfancy}{}{g (\zvector)}{d\qhatn(\zvector)}  -  \integral{\zfancy}{}{g (\zvector)}{dp(\zvector)}  \right]          \\
    & =  \ktheta W_1(\qhatn(\zvector), p(\zvector))  ,
\end{split}
\end{equation*}
where $\gfancy_\theta = \theset{g: \zfancy \rightarrow \real \,\text{ s.t. } g=f\circ g_\theta \text{ and } f\in \lipschitz{1}(\xfancy, \real)}$ and the inequality holds because $\gfancy_\theta \subseteq \lipschitz{\ktheta}(\zfancy, \real)$, since $g_\theta: \zfancy\rightarrow \xfancy$ is $\ktheta$-Lipschitz. Now, since $(p, q) \mapsto W_1(p, q)$ is convex, the definition of $\qhatn(\zvector)$ implies
\begin{equation}\label{eq-w1-w2-qhat}
 W_1(\qhatn(\zvector), p(\zvector))  \leq \unsur{n}\somme{i=1}{n}{W_1(\qphi{\zgivenx_i}, p(\zvector))} \leq 
 \unsur{n}\somme{i=1}{n}{W_2(\qphi{\zgivenx_i}, p(\zvector))}.
\end{equation}
Since, by Equation \eqref{eq-w2-gaussian-diag},
\[
W_2(\qphi{\zgivenx_i}, p(\zvector))^2 = \norme{\muphi{\xvector_i}}^2 + \norme{\sigmaphi{\xvector_i} - \unvec}^2, 
\]
we obtain
\[
W_1(\muhatn, \pushf{g_\theta }{p(\zvector)}) \leq \frac{\ktheta}{n} \somme{i=1}{n}{ \sqrt{ \norme{\muphi{\xvector_i}}^2 + \norme{\sigmaphi{\xvector_i} - \unvec}^2} }.
\]
\end{proof}

\subsection{Proof of Theorem \ref{thm-regen-bounded}}

\begin{proof}[Proof of Theorem \ref{thm-regen-bounded}]

Recall from Lemma \ref{lem-mu-muhatn} that with probability at least $1-\delta$ over the random draw of $S\sampled \mu\otimesn$,
\begin{equation}\label{eq-app-someineq}
\begin{split}
\lambda W_1(\mu, \muhatn) \leq \frac{\lambda}{n} \somme{i=1}{n}{ \left( \expectseul{\zvector \sampled q(\zvector|\xvector_i)}{\lossrec(\zvector, \xvector_i)}  \right)    } 
  + \somme{i=1}{n}{\kl{q(\zvector|\xvector_i)}{p(\zvector)}  }  
+  \\
\log \unsur{\delta} + \log \expectseul{S\sampled \mu\otimesn}{   \expectseul{\zvector\sampled p(\zvector)}{\expon{ \lambda  \left( \expect{\xvector\sampled \mu}{\lossrec(\zvector, \xvector)} -  \unsur{n} \somme{i=1}{n}{ \lossrec(\zvector, \xvector_i)}      \right)  }    }        }.   
\end{split}
\end{equation}

In order to prove Theorem  \ref{thm-rec-bounded} in section \ref{sec-proof-thm-rec-bounded}, we proved that
\[ 
\expectseul{S\sampled \mu\otimesn}{  \exponbig{\lambda \left( \expect{\xvector\sampled \mu}{\lossrec(\zvector, \xvector)} - \unsur{n} \somme{i=1}{n}{\lossrec(\zvector, \xvector_i)}       \right)}      } 
\leq 
\exponbig{\frac{\lambda^2 \Delta^2}{8n}}.
\]
Now, we can reuse this inequality to upper-bound the last term on the right-hand side of Equation~\eqref{eq-app-someineq}. We obtain the desired theorem:
under the assumptions of Theorem \ref{thm-rec-bounded}, with probability at least $1-\delta$ over the random draw of $S\sampled \mu\otimesn$, the following holds for any posterior $\qphi{\zgivenx}$:
\begin{equation*}
\begin{split}
W_1(\mu, \muhatn) \leq 
\frac{1}{n} \somme{i=1}{n}{ \left\{ \expectseul{ \qphi{\zvector|\xvector_i}   }{\lossrec(\zvector, \xvector_i)}  \right\}    }  
  + 
\unsur{\lambda} \left( \somme{i=1}{n}{\kl{\qphi{\zvector|\xvector_i}}{p(\zvector)}  }  + 
\log \unsur{\delta} + \frac{\lambda^2 \Delta^2}{8n}   \right).
\end{split}
\end{equation*}

\end{proof}

\subsection{Proof of Theorem \ref{thm-gen-bounded}}
\begin{proof}[Proof of Theorem \ref{thm-gen-bounded}]
Theorem \ref{thm-gen-bounded} is a direct consequence of Theorem \ref{thm-regen-bounded} and Lemma \ref{lem-muhatn-gpz} applied to Equation \eqref{eq-app-triangle-mu}.
\end{proof}

\subsection{Proof of Theorem \ref{thm-regen-manif}}
\begin{proof}[Proof of Theorem \ref{thm-regen-manif}]
Recall from Lemma \ref{lem-mu-muhatn} that with probability at least $1-\delta$ over the random draw of $S\sampled \mu\otimesn$,
\begin{equation*}
\begin{split}
\lambda W_1(\mu, \muhatn) \leq \frac{\lambda}{n} \somme{i=1}{n}{ \left( \expectseul{\zvector \sampled q(\zvector|\xvector_i)}{\lossrec(\zvector, \xvector_i)}  \right)    } 
  + \somme{i=1}{n}{\kl{q(\zvector|\xvector_i)}{p(\zvector)}  }  
+  \\
\log \unsur{\delta} + \log \expectseul{S\sampled \mu\otimesn}{   \expectseul{\zvector\sampled p(\zvector)}{\expon{ \lambda  \left( \expect{\xvector\sampled \mu}{\lossrec(\zvector, \xvector)} -  \unsur{n} \somme{i=1}{n}{ \lossrec(\zvector, \xvector_i)}      \right)  }    }        }.   
\end{split}
\end{equation*}

We can then use Lemma \ref{lemma:exp-moment-thm-rec-manif} which stated that
    \begin{equation}
\log \expectseul{\zvector\sampled p(\zvector)}{
\expectseul{S\sampled \mu\otimesn}{  \exponbig{\lambda \left( \expect{\xvector\sampled \mu}{\lossrec(\zvector, \xvector)} - \unsur{n} \somme{i=1}{n}{\lossrec(\zvector, \xvector_i)}  \right)} } } \leq 
\frac{\lambda^2 \ketoil^2}{2n}.
\end{equation}
The expectations over $\zvector$ and $S$ can be swapped using Fubini's Theorem. Hence, combining Lemma~\ref{lemma:exp-moment-thm-rec-manif} and Lemma~\ref{lem-mu-muhatn}, we obtain Theorem \ref{thm-regen-manif}: with probability at least $1 -\delta$ over the random draw of $S\sampled \mu\otimesn$, the following holds for any posterior $\qphi{\zgivenx}$.
\begin{equation*}
\begin{split}
W_1(\mu, \muhatn) \leq
 \frac{1}{n} \somme{i=1}{n}{ \left\{ \expectseul{ \qphi{\zvector|\xvector_i}   }{\lossrec(\zvector, \xvector_i)}  \right\}    }  
  + \unsur{\lambda} \left( 
  \somme{i=1}{n}{\kl{\qphi{\zvector|\xvector_i}}{p(\zvector)}  }  + 
\log \unsur{\delta} + \frac{\lambda^2 \ketoil^2}{2n} \right).
\end{split}
\end{equation*}
\end{proof}

\subsection{Proof of Theorem \ref{thm-gen-manif}}
\begin{proof}[Proof of Theorem \ref{thm-gen-manif}]
Theorem \ref{thm-gen-manif} is a direct consequence of Theorem \ref{thm-regen-manif} and Lemma \ref{lem-muhatn-gpz} applied to Equation \eqref{eq-app-triangle-mu}.
\end{proof}

\section{Numerical Experiments}
\begin{figure}
\begin{center}
\includegraphics[width=.8\linewidth, height=4cm]{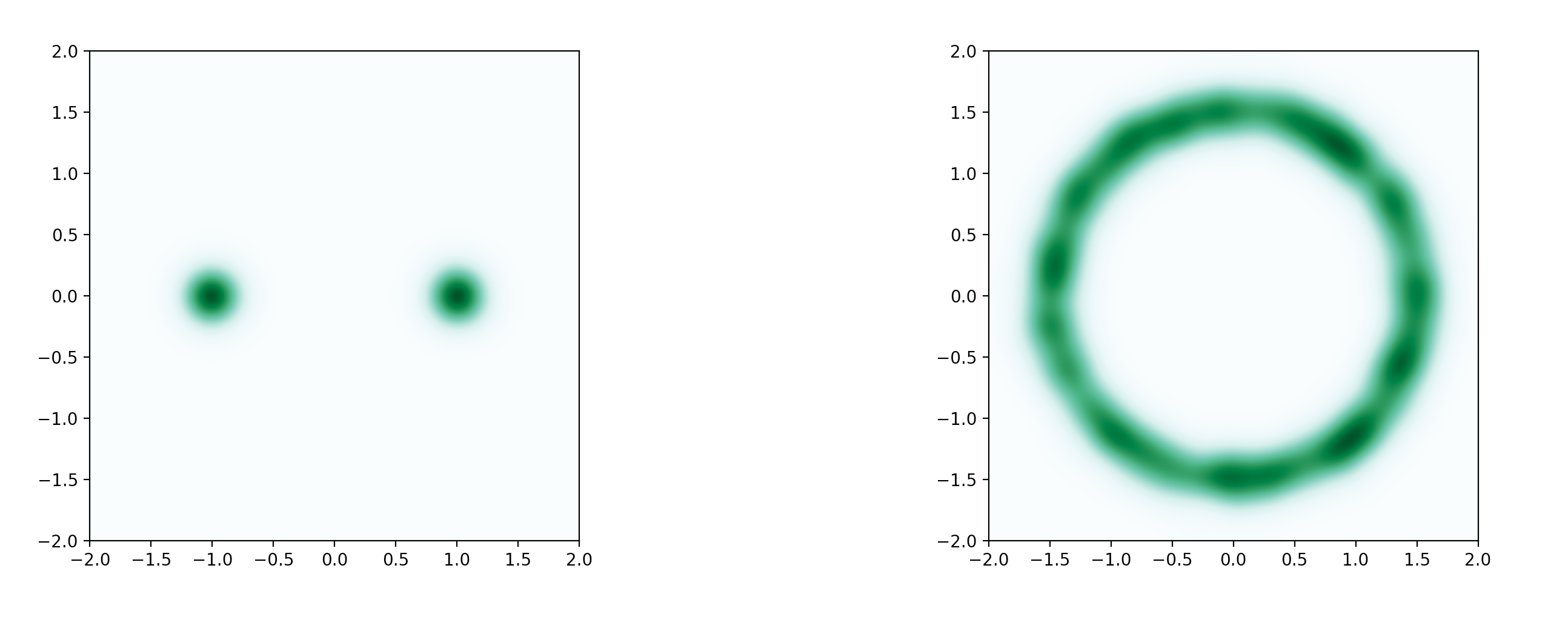}
\end{center}
\caption{Samples from the real datasets}
\label{fig-vae-real-data}
\end{figure}

We computed the numerical value of the bound of Theorem~\ref{thm-rec-bounded}. 
We performed the experiments on two $2$-dimensional synthetic datasets. The first one is a mixture of two isotropic Gaussian distributions on $\real^2$ centered at $(-1, 0)$ and $(1, 0)$ respectively, and with standard deviation $\sigma =0.1$ and null covariances. The second dataset consists of noisy samples arranged in a circle centered at the origin, with radius $1.5$ and standard deviation $\sigma=0.1$. Both datasets are truncated so that no sample is over 4 standard deviations away from its corresponding mean. This is to formally ensure that the diameter of the instance spaces is finite, as required by Theorem~\ref{thm-rec-bounded}. The sizes of the training, validation and test sets are respectively 50,000, 20,000 and 20,000. Samples from the two datasets are shown in Figure~\ref{fig-vae-real-data}.

\begin{figure}
\begin{center}
\includegraphics[width=.95\linewidth, height=2.5cm]{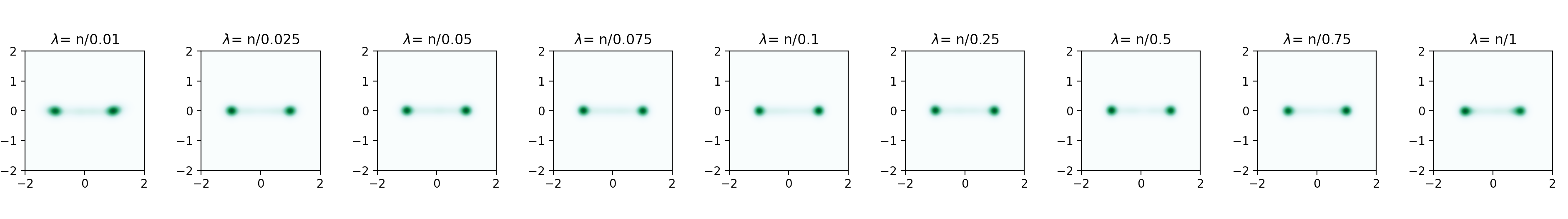}
\vspace{0.5cm}
\includegraphics[width=.95\linewidth, height=2.5cm]{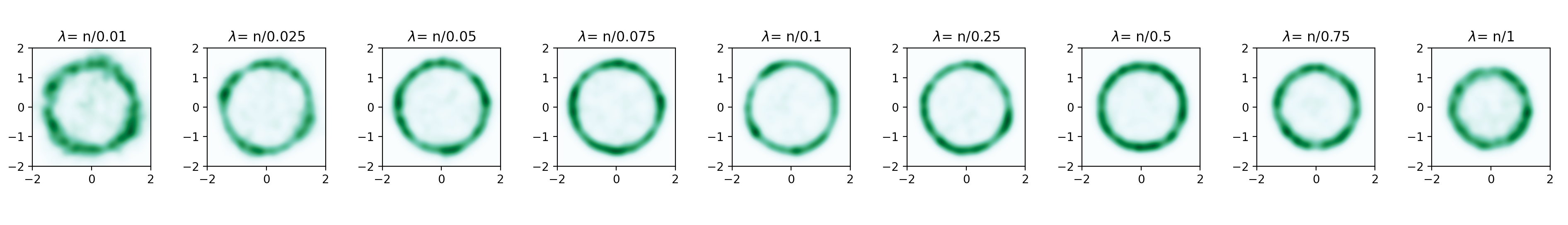}
\end{center}
\vspace{-0.8cm}
\caption{Samples from the models trained on the 2-Gaussian dataset (top) and the Circle dataset (bottom).}
\label{fig-vae-fake}
\end{figure}

We used the same architecture and hyperparameters for both datasets. The encoder and decoder are fully connected networks with 3 hidden layers and 100 hidden units per layer. We also set the Lipschitz constants of the encoder and decoder networks to $\kphi = \ktheta = 2$. In order to enforce Lipschitz continuity, we used Björk orthonormalization \citep{bjorck} with GroupSort activations \citep{sorting-out}, and we utilized the implementation of Lipschitz layers by \citet{sorting-out}. Note that \citet{cert-robust-vaes} performed experiments with VAEs with fixed Lipschitz constants, but we did not directly use their implementation because of a difference in the definition of the Lipschitz norm of the encoder, which affects the implementation. Note also that unlike the usual computations of PAC-Bayesian bounds \citep{perez-ortiz}, our implementation does not use probabilistic neural networks. It uses deterministic networks, as it is usual for VAEs, because our analysis did not include additional stochasticity. We used the MSE as the reconstruction loss during training, and computed the bounds on validation datasets. The samples from the different models are displayed in Figure~\ref{fig-vae-fake}.

Recall the inequality of Theorem~\ref{thm-rec-bounded}:
\begin{equation}\label{eq-recall-rec-bounded}
\begin{split}
 \underbrace{\expectseul{\xvector\sampled \mu }{\expectseul{\qphi{\zvector|\xvector}  }{\lossrec(\zvector, \xvector)}} }_\text{Test Rec. Loss}     
 \leq
 \underbrace{\frac{1}{n} \somme{i=1}{n}{ \left\{ \expectseul{ \qphi{\zvector|\xvector_i} }{\lossrec(\zvector, \xvector_i)}  \right\}  }  }_\text{Emp. Rec. Loss}
  + \underbrace{\unsur{\lambda} \somme{i=1}{n}{\kl{\qphi{\zvector|\xvector_i}}{p(\zvector)}  }}_\text{Emp. KL loss}  +   \\ 
   \underbrace{\kphi \ktheta \Delta}_\text{Avg distance}   
+ \underbrace{\frac{\lambda \Delta^2}{8n}}_\text{Exp. moment}  + \unsur{\lambda}\log \unsur{\delta} .  
\end{split}
\end{equation}

Tables \ref{tab-vae-g} and \ref{tab-vae-c} show the numerical values of the bound of Theorem~\ref{thm-rec-bounded} for different values of $\lambda$. The first column is approximated using the test set, and the last one refers to all the right-hand side of \eqref{eq-recall-rec-bounded}. The empirical reconstruction and KL losses are computed using the validation set, since, as mentioned in the main paper, the bounds need to be computed using a set independent from the training set.

\begin{table}
\begin{center}
\begin{tabular}{|c|c|c|c|c|c|c|}
\hline
$\lambda$ & \textbf{Test Rec. loss }& Emp. Rec. loss & Emp. KL loss  &  Exp. moment & \textbf{Bound} \\  \hline
$n/0.01$    &     0.1107 &     0.1110     & 0.0192 &         89.00 &       99.80 \\ \hline
$n/0.025$ &     0.1228 &     0.1237     & 0.0505 &       35.60 &       46.45 \\ \hline
$n/0.05$   &     0.1299 &     0.1299     & 0.1010 &          17.80 &       28.70  \\ \hline
$n/0.075$ &     0.1388 &     0.1403     & 0.1511 &          11.867 &       22.83  \\ \hline
$n/0.1$      &     0.1425 &     0.1436     & 0.2003 &        8.900 &      19.92 \\ \hline
$n/0.25$  &      0.1707 &     0.1732     & 0.4883 &         3.560 &       14.89   \\ \hline
$n/0.5$    &      0.2120 &     0.2162     & 0.9602 &          1.780 &      13.63  \\ \hline
$n/0.75$   &     0.2718 &     0.2725     & 1.4122 &         1.1868 &       13.54  \\ \hline
$n/1$      &        0.3586 &     0.3596     & 1.8593 &        0.8901 &       13.78  \\ \hline
\end{tabular}
\end{center}
\caption{Table showing the values of the different quantities of Equation~\ref{eq-recall-rec-bounded} for the \guillemets{2-Gaussian} dataset. The upper bound on the average distance term is $10.67$.}
\label{tab-vae-g}
\end{table}

\begin{table}
\begin{center}
\begin{tabular}{|c|c|c|c|c|c|c|}
\hline
$\lambda$ & \textbf{Test Rec. loss }& Emp. Rec. loss & Emp. KL loss  &  Exp. moment & \textbf{Bound} \\  \hline
$n/0.01$ &       0.095 &        0.0959 &      0.0197 &         180.50 &           195.81  \\ \hline
$n/0.025$ &     0.1354 &      0.1362 &      0.0525 &        72.20 &           87.59 \\ \hline
$n/0.05$ &       0.1785 &      0.1783 &      0.1058 &         36.10 &           51.58  \\ \hline
$n/0.075$ &     0.2005 &      0.2020 &      0.1587 &           24.07 &           39.63  \\ \hline
$n/0.1$ &         0.2245 &      0.2247 &      0.2117 &         18.05 &           33.69 \\ \hline
$n/0.25$ &      0.3498 &      0.3486 &      0.5160 &            7.220 &           23.28   \\ \hline
$n/0.5$ &         0.5026 &      0.4940 &      0.9997 &        3.610 &           20.30  \\ \hline
$n/0.75$ &      0.6171 &      0.6154 &      1.4691 &        2.406 &           19.691  \\ \hline
$n/1$ &            0.7513 &      0.7499 &      1.9314 &       1.805 &           19.686  \\ \hline
\end{tabular}
\end{center}
\caption{Table showing the values of the different quantities of Equation~\ref{eq-recall-rec-bounded} for the \guillemets{Circle} dataset. The upper bound on the average distance term is $15.2$.}
\label{tab-vae-c}
\end{table}

From Tables \ref{tab-vae-g} and \ref{tab-vae-c}, once can see that the bounds are dominated by two terms: the average distance and the exponential moment. Although as $\lambda$ approaches $n$, the exponential moment gets smaller and the main influence comes from the upper bound on the average distance. Hence, in order to tighten the bound, one may need to derive tighter upper bounds on the average distance, or derive versions of Theorem~\ref{thm-rec-bounded} where this term is replaced by a numerically smaller one.


\vspace{2cm}
\section{Additional Results and Remarks}
This section contains additional remarks and discussions. We start with possible extensions of our results.

\subsection{The variance of the likelihood}

Our definition of the decoder network's output (the function $\gthetaseul : \zfancy \rightarrow \xfancy$) only considers the deterministic part of the decoder. In other words, our results only apply to VAEs whose likelihood has constant variance. However, they can be extended to cases when the variance of the likelihood is optimized, but at a cost. We discuss separately the two cases where the variance depends on individual datapoints or not.

\paragraph{Instance-independent variance.} 
If the standard deviation $\sigma$ of the decoder is fixed, then we have $\sigma \propto \frac{n}{\lambda}$, (recall the hyperparameter $\lambda$ from Theorem~\ref{thm-after-ipm-assum} and subsequent theorems). Hence, optimizing $\sigma$ corresponds to optimizing $\lambda$, which is non-trivial in PAC-Bayes. Indeed, most PAC-Bayes bounds (including ours) do not directly allow one to optimize $\lambda$ (see Section 2.1.4 of \citet{friendly}). Although there are some ways around this restriction, we are not aware of any results that allow one to optimize in the general case (meaning continuous values of $\lambda$ and unbounded loss). For $[0, 1]$-bounded loss functions, \citet{thieman} developed a PAC-Bayes bound uniformly valid for a trade-off parameter $\lambda'$, and show that one can optimize w.r.t. both the posterior and $\lambda'$, under certain assumptions. For unbounded losses, if one assumes $\lambda \in \Lambda$, where $\absolu{\Lambda}$ is finite, a union bound argument allows one to make the bound uniform with respect to $\lambda$, at the cost of $\log \absolu{\Lambda}$ (see \citet{friendly}). One can still optimize with respect to a continuous set $\Lambda$, by considering a grid. For instance, if one considers $\Lambda \cap \theset{1, \dots, n}$, then the penalty is $\log n$ and if one considers $\Lambda \cap \theset{\expon{k}: 1 \leq k \leq n}$, the penalty is $\log \log n$.

\paragraph{Instance-dependent variance.} 
Now, assume the standard deviation is dependent on individual instances. Say we define the reconstruction loss as $\loss_\theta(\zvector, \xvector) = \unsur{\sigma_\theta(\zvector)} \norm{\xvector - \gtheta{\zvector}}$, where $\sigma_\theta : \zfancy \rightarrow \real_{> 0}$. Because of the division by $\sigma_\theta(\zvector)$, let us assume that there is a fixed upper bound $\sigma_1 > 0$ such that $\sigma_\theta(\zvector) > \sigma_1$, for any $\zvector \in \zfancy$. There are two main tasks: making sure Assumption~\ref{assum-ipm} is satisfied, and bounding the exponential moment of Theorem \ref{thm-gen-vae}, with this new loss function. 

Verifying Assumption~\ref{assum-ipm} is equivalent to showing that Proposition~\ref{prop-assum-ipm} is verified for this new loss function $\loss_\theta$. The second part of the proof of Proposition 4.1 tells us that we need to show that $\loss_\theta$ is Lipschitz-continuous. Note that in general, the product of real-valued Lipschitz functions is not Lipschitz. Hence, we assume, in addition, that $\norm{\xvector - \gtheta{\zvector}} \leq M < \infty$. The following proposition shows that Assumption \ref{assum-ipm} is satisfied with the constant $K = K_\phi \left( \frac{K_\sigma M}{\sigma_1^2} + \frac{K_\theta}{\sigma_1} \right)$.

\begin{proposition}
Consider a VAE with parameters $\phi$ and $\theta$ and let $\kphi, \ktheta \in \real$ be the Lipschitz norms of the encoder and decoder respectively. Also, consider the loss function $\lrec : \zfancy \times \xfancy \rightarrow \real$ defined as
\[
\lrec(\zvector, \xvector) = \unsur{\sigma_\theta(\zvector)} \norme{\xvector - \gtheta{\zvector}}
\]
where $\sigma_\theta : \zfancy \rightarrow \real_{>0}$ is $K_\sigma$-Lipschitz. Assume and for all $\zvector \in \zfancy$, $\sigma_\theta(\zvector) > \sigma_1$ and $\norme{\xvector - \gtheta{\zvector}} \leq M$ for some fixed $0 < \sigma_1 < 1$ and $M > 0$. Then the variational distribution $\qphi{\zgivenx}$ satisfies Assumption \ref{assum-ipm} with  $\efancy  = \theset{f: \zfancy \rightarrow \real : \norme{f}_\textrm{Lip} \leq \frac{K_\sigma M}{\sigma_1^2} + \frac{K_\theta}{\sigma_1}}$, $K = \kphi \left( \frac{K_\sigma M}{\sigma_1^2} + \frac{K_\theta}{\sigma_1} \right)$, and $\loss = \lrec$.
\end{proposition}

\begin{proof}
The first part of Assumption~\ref{assum-ipm} is satisfied, since $\frac{K_\sigma M}{\sigma_1^2} + \frac{K_\theta}{\sigma_1} > \ktheta$. Now, for the second part of Assumption~\ref{assum-ipm}, we need to show that $\lrec$ is $\frac{K_\sigma M}{\sigma_1^2} + \frac{K_\theta}{\sigma_1}$-Lipschitz continuous. First, 
\[
\absolu{ \unsur{\sigma_\theta(\zvector_1)} - \unsur{\sigma_\theta(\zvector_2)} } = \absolu{\frac{\sigma_\theta(\zvector_2) - \sigma_\theta(\zvector_1)}{\sigma_\theta(\zvector_1) \sigma_\theta(\zvector_2)}} \leq \frac{K_\sigma \norme{\zvector_1 - \zvector_2}}{\sigma_1^2}.
\]

We have
\begin{align*}
\setlength{\jot}{10pt}
\begin{split}
\absolu{\lrec(\zvector_1, \xvector) - \lrec(\zvector_2, \xvector  } 
    & =  \absolu{\unsur{\sigma_\theta(\zvector_1)} \norme{\xvector - \gtheta{\zvector_1}} - \unsur{\sigma_\theta(\zvector_2)} \norme{\xvector - \gtheta{\zvector_2}}    }      \\
    & =  \absolu{\unsur{\sigma_\theta(\zvector_1)} - \unsur{\sigma_\theta(\zvector_2) }  } \norme{\xvector - \gtheta{\zvector_1}}  + 
    \unsur{\sigma_\theta(\zvector_2) } \absolu{ \norme{\xvector - \gtheta{\zvector_1}} -  \norme{\xvector - \gtheta{\zvector_2}}    }           \\
    & \leq \frac{K_\sigma M}{\sigma_1^2} \norme{\zvector_1 - \zvector_2}  + \frac{\ktheta}{\sigma_1} \norme{\zvector_1 - \zvector_2}          \\
    & = \left( \frac{K_\sigma M}{\sigma_1^2} + \frac{K_\theta}{\sigma_1} \right)    \norme{\zvector_1 - \zvector_2}     \\
\end{split}
\end{align*}
\end{proof}

Now, let us focus on bounding the exponential moment. In this case, when the instance space is bounded, the upper bound on the exponential moment (in the proof of Theorem 4.3) is:
    \[
    \frac{\lambda^2 \Delta^2}{8n \sigma_1^2}, \quad \text{ instead of } \quad  \frac{\lambda^2 \Delta^2}{8n}.
    \]
    And under the manifold assumption, we get the following upper bound (in the proof of Theorem 4.4):
    \[
    \frac{\lambda^2 K_*^2}{2n \sigma_1^2}, \quad \text{ instead of } \quad \frac{\lambda^2 K_*^{2}}{2n}
    \]
Note that although the upper bounds on the average distance remain unchanged, the coefficient $K_\phi K_\theta$ is replaced by $K_\phi \left( \frac{K_\sigma M}{\sigma_1^2} + \frac{K_\theta}{\sigma_1} \right)$, which is larger, specially if $\sigma_1$ is very small.

\subsection{Uniformity with respect to $\theta$}
As mentioned in the main paper, although our bounds hold uniformly for any encoder $\phi$, they only hold for a given decoder $\theta$. 
the consequence of this limitation is that the numerical computations of the bounds need to be done on a sample set disjoint from the training set (e.g. a validation or test set). Let $\Theta$ denote a set of decoder parameters over which the optimization is performed.

From a theoretical perspective, the union bound can be used to circumvent this issue, when we consider a finite set of parameters $\Theta$. In that case, the $\log \unsur{\delta}$ in Theorem~\ref{thm-after-ipm-assum} becomes $\log \frac{\absolu{\Theta}}{\delta}$, which loosens the bound. Moreover, since $\Theta$ denotes a set of neural network parameters, this assumption may not be appropriate unless one chooses a very large set $\Theta$, which can significantly loosen the bound.

Another option would be to make assumptions on the complexity of the set of loss functions $\theset{\lossrec : \theta \in \Theta}$ parameterized by decoder parameters $\theta \in \Theta$  (e.g. the Rademacher complexity), in order to obtain uniform bounds in a more general case. We leave such explorations to future works.

\subsection{Additional Remarks}

\begin{remark}[Alternate formulation of Assumption \ref{assum-ipm}]\label{rem-assum-alternate}
We can provide an equivalent formulation of Assumption \ref{assum-ipm}. A posterior $q(h|\xvector)$ and a loss function $\loss$ satisfy Assumption~\ref{assum-ipm} with a constant $K>0$ if and only if for any $\xvector \in \xfancy$,
\[
\absolu{ \expectseul{h\sampled q(h|\xvector_1)}{\loss(h, \xvector)} - \expectseul{h\sampled q(h|\xvector_2)}{\loss(h, \xvector)}  } \leq Kd(\xvector_1, \xvector_2) .
\]
The formulation given in the paper is more intuitive, but this expression shows that the specific choice of $\newf $ does not matter. The equivalence of the two formulations is a consequence of the definition of an IPM.
\end{remark}

\begin{remark}[Prior Learning in PAC-Bayes]\label{rem-prior-learning}
The majority of PAC-Bayesian bounds \citep{some_pb_thms, seeger2002, pac-bayes-linear, pb-gen-models} require the prior distribution $p$ on the hypothesis class to be independent of the training set\footnote{PAC-Bayesian bounds with data-dependent priors were developed by \citet{diff-privacy, beyond-usual}.}. In practice, this means one has to use data-free priors when minimizing PAC-Bayes bounds. Since, in that case, the learned posterior is likely very far from the prior, the KL-divergence tends to be orders of magnitude larger than the empirical risk. In practice, this means the optimization is monopolized by the KL-divergence, leading to a poor performance of the learning algorithm. In order to avoid this issue and still obtain a valid certificate, the following \guillemets{prior learning trick} is used. Split the training set $S=\theset{\listen{\xvector}}$ in two disjoint subsets $S_1, S_2$, where $\absolu{S_1} = n_0, \absolu{S_2} = n-n_0$ with $n_0 < n$. Then, learn the prior $p$ on $S_1$, learn the posterior $q$ on $S$ (the whole training set), and compute the certificate on $S_2$. 

The reason why this trick cannot be directly applied to circumvent the fact that our bounds are valid for a given decoder, is that the encoder and the decoder are jointly optimized in VAEs. Hence, one has to make sure the samples used to learn the encoder (hence, train the model) are not used in the computation of the risk certificate. We emphasize that in our case, the issue does not lie in the learning of the prior (the standard VAE considers a standard Gaussian prior), but of the loss function $\lossrec$, which is dependent on the decoder's parameters $\theta$.

\end{remark}

\end{document}